\documentclass{article}
\usepackage{ijcai17}

\usepackage{cite}
\usepackage{times}
\usepackage{times}
\usepackage{latexsym}
\usepackage{xspace}
\usepackage{color}
\usepackage{url}
\usepackage{graphicx}
\usepackage{multirow}
\usepackage{makecell}
\usepackage{amsmath}
\usepackage{amssymb}
\usepackage[
  bookmarks=false,
  pdfpagelabels=false,
  hyperfootnotes=false,
  hyperindex=false,
  pageanchor=false,
  colorlinks,
]{hyperref}
\usepackage{subcaption}
\usepackage{bm}
\usepackage{algorithm} 
\usepackage[noend]{algpseudocode}
\usepackage{mathtools}
\mathtoolsset{showonlyrefs=true}
\usepackage{enumitem}

\title{Creatism: A deep-learning photographer capable of creating professional work}

\author{Hui Fang \\ \small hfang@google.com \normalsize \\ \small Google 
\And Meng Zhang \\ \small zhangmeng@google.com\normalsize \\ \small Google}

\makeatletter
\renewcommand{\paragraph}{%
  \@startsection{paragraph}{4}%
  {\z@}{1ex \@plus 1ex \@minus .2ex}{-1em}%
  {\normalfont\normalsize\bfseries}%
}
\makeatother

\begin{document}

\maketitle

\begin{abstract}
Machine-learning excels in many areas with well-defined goals.
However, a clear goal is usually not available in art forms, such as photography. 
The success of a photograph is measured by its aesthetic value,
a very subjective concept.
This adds to the challenge for a machine learning approach.

We introduce Creatism, a deep-learning system for artistic content creation.
In our system, we break down aesthetics into multiple aspects,
each can be learned individually from a shared dataset of professional examples.
Each aspect corresponds to an image operation that can be optimized efficiently.
A novel editing tool, dramatic mask, is introduced as one operation that improves dramatic 
lighting for a photo.
Our training does not require a dataset with before/after image pairs, or any additional 
labels to indicate different aspects in aesthetics.

Using our system, we mimic the workflow of a landscape photographer,
from framing for the best composition to carrying out various post-processing operations.
The environment for our virtual photographer is simulated by a collection of panorama images from Google Street View.
We design a ``Turing-test"-like experiment to objectively measure quality of its creations,
where professional photographers rate a mixture of photographs from different sources blindly.
Experiments show that a portion of our robot's creation can be confused with professional work.

\end{abstract}

\section{Introduction}

Great progress has been made in both camera hardware and computational photography,
such that a modern cell phone can take technically solid photographs, 
in terms of exposure level, noise level, pixel sharpness, color accuracy, etc.
However, a good photo should be not only technically solid, but also aesthetically pleasing.

Aesthetics is vague and subjective, a metric hard to define scientifically.
Multiple research exists ~\cite{6247954} ~\cite{DBLP:journals/corr/KongSLMF16} 
to collect dataset to define aesthetic quality.
Generating images towards top aesthetic quality is an even harder problem.
A naive approach using a single aesthetic prediction is insufficient to capture different
aspects in aesthetics, as we will show in experiment.

In this paper, we introduce Creatism, a deep-learning system for artistic content creation.
Here aesthetics is treated not as a single quantity, but as a combination of different aspects.
Each aspect is learned independently from professional examples,
and is coupled with an image operation that can modify this aspect.
By making image operations semi-orthogonal, we can efficiently optimize a photo one aspect at a time.

Another advantage of coupling an aesthetic aspect with an image operation is that we can
simulate ``negative" examples tailored towards that aspect. 
This gets rid of the need to collect before/after image pairs from professionals to indicate
how to improve each aspect.
In this project the dataset for aesthetic aspects training is a collection of professional-quality
photos with no additional labels.

In addition to learn aesthetic aspects with known image operations,
we show that it is also possible to define new operations from this unlabeled dataset.
By combining a set of existing image filters to generate negative examples, 
we train a new image operation, dramatic mask, that can enhance dramatic lighting in photos.

One standing problem for current enhancement works is quality metric,
especially on higher end of aesthetics.
A user-study with image comparison tells which images are better.
But a ``better" image may still remain mediocre.
A direct scoring method is limited by the expertise of the evaluators, who are also susceptive
to bias in a non-blind setting.

What is the ``ultimate" metric for the highest aesthetic standard human can define?
Drawing inspiration from the famous Turing-test, we propose the following metric:
\begin{itemize}
\item A generated photo is at professional level, if professionals, in a blind test, can not tell whether it is created by an algorithm, or by another professional.
\end{itemize}

In our work we use ``professional", i.e. professional photographers, to represent best experts
in photography.
The standard can be further raised by replacing that word with, say, ``top 10 master photographers in the world".

We work with professional photographers to define 4 levels of aesthetic
quality, with the top level ``pro".
In experiment we ask professionals to rate a random mixture of photos into different levels, with photos
taken by our robot mixed in.
For robot creations with high prediction scores, about $40\%$ ratings we receive are at semi-pro to pro level.
While we didn't beat our ``Turing-test" consistently, 
we show that creating photos using machine learning at professional quality is made possible.

Our contributions in this paper are:
\begin{itemize}
\item A deep-learning system that learns different aspects of aesthetics 
and applies image operations to improve each aspect in a semi-orthogonal way, 
using a dataset of professional quality photos with no additional labels,
or before/after image pairs.
\item Introduce dramatic mask as a novel image operation to enhance dramatic lighting in photos.
\item In a ``Turing-test"-like experiment, we show that our system can create photos from 
environment with some of them at semi-pro to pro level.
\end{itemize}

The rest of the paper is organized as following: 
Related works are discussed in Section \ref{sec:related_work}.
We describe the framework of Creatism system in Section \ref{sec:problem_formation},
including how to train deep-learning models for different aspects in aesthetics, 
and how to optimize each aspect independently in a photo using its associated image operation.
Two special image operations, cropping and dramatic mask, are discussed in Section \ref{sec:special_ops}.
To evaluate photos created by our robot, aesthetic levels are defined in Section \ref{sec:scores}.
Experiments are presented in Section \ref{sec:experiments}.

\section{Related Work} \label{sec:related_work}

Datasets were created that contain ratings of photographs based on
aesthetic quality ~\cite{6247954} ~\cite{DBLP:journals/corr/KongSLMF16} ~\cite{7243357}.
Machine-learning methods have been used to train models on aesthetic quality using
such datasets, with either hand-crafted features ~\cite{6247954} ~\cite{DBLP:journals/corr/MarchesottiMP14},
or deep neural network ~\cite{Lu:2014:RRP:2647868.2654927} ~\cite{7410476}.

Image enhancement can be guided by learned aesthetic model.
In ~\cite{YanZW+16} color transformation is predicted from
pixel color and semantic information, learning from before and after
image pairs edited by a photographer.

Beyond judging a photo as simply ``good" or ``bad", multiple aspects of aesthetics can
be learned to control image filters.
In ~\cite{acceptable}, images with brightness and contrast variations were evaluated as
acceptable or not by crowd computing, which in turn predicts acceptable range of these
filters on new images.
However, sampling images in $2-D$ space of brightness and contrast and having them labeled
by people does not scale well.
In this paper we present an approach that studies each filter in its own subspace.
We also get rid of human labelling in this step.

AVA dataset was used to train image cropping in ~\cite{7780429},
where a sliding-window was applied to an image to yield the best crop.
A recent approach in ~\cite{DBLP:journals/corr/ChenKSCM17} is similar in spirit to ours, 
where the authors learn a model for aesthetic composition by using random crops as negative examples.
Our approach decouples multiple aesthetic aspects including composition. 
In the special case of cropping, we show that a hybrid approach leads to
more variations of good cropping candidates.

Pair-wise image style transfer is another way to enhance images.
Deep-learning approaches, pioneered by ~\cite{DBLP:journals/corr/GatysEB15a},
show huge advantage over
traditional texture-synthesis based approaches ~\cite{Hertzmann:2001:IA:383259.383295}. 
Recent research also transfers styles from photos to photos while preserving realistic details
in results ~\cite{DBLP:journals/corr/LuanPSB17}.
However, such transfer requires the user to manually provide an image or painting as the target. 
The success of transfer heavily depends on how suitable the target is. 

In Generative-Adversarial Networks (GANs) ~\cite{NIPS2014_5423},
a generative model G and a discriminative model D are simultaneously trained.
It leads to amazing content creation ability, being able to generate plausible images
of different categories ~\cite{DBLP:journals/corr/RadfordMC15}
~\cite{DBLP:journals/corr/NguyenYBDC16}
~\cite{DBLP:journals/corr/ZhuPIE17}.
Conditions can be introduced to control GAN's results ~\cite{DBLP:journals/corr/MirzaO14}.
Such condition may come from another image, such that generated results become a variation
of the input image ~\cite{DBLP:journals/corr/IsolaZZE16}.

However, GAN results usually contain noticeable artifacts.
This limits its application in image enhancement. 
In this paper we introduce an image operation ``dramatic mask" that
uses a similar structure as DC-GAN~\cite{DBLP:journals/corr/IsolaZZE16}.
Instead of outputting pixels directly, it creates a low-resolution mask to modulate brightness,
conditioned by original image.
This operation enhances dramatic lighting in many photos.

Evaluation on aesthetics is very subjective.
Comparing to ground truth ~\cite{7780429}  ~\cite{DBLP:journals/corr/ChenKSCM17} 
is only viable when ground-truth exists.
User study is another option, where typically 2 or more images are compared to
yield the best one ~\cite{YanZW+16}.
However, the winning image in a group may still be of low aesthetic quality.
Previous study shows that image comparison can produce absolute ranking given enough pairs
~\cite{Mantiuk:2012:CFS:2393476.2393485}.
However collecting huge amount of pairs from professionals is not practical.
Instead, we adapt an approach with an absolute scoring system.
Our evaluation is a ``Turing-test" on aesthetic quality,
where professionals are asked to give scores to a mixture of photos with predefined absolute meaning.

\section{Problem Formulation} \label{sec:problem_formation}

Assume there exists a universal aesthetics metric, $\Phi$, that gives a higher score
for a photo with higher aesthetic quality.
A robot is assigned a task to produce the best photograph $P$, measured by $\Phi$, from its environment $\mathcal{E}$.
The robot seeks to maximize $\Phi$ by creating $P$ with optimal actions, controlled by parameters $\{x\}$:

\begin{equation}
\label{eq:objective}
\underset{\{x\}}{\arg\max} ( \Phi(P(\mathcal{E}, \{x\})))
\end{equation}

The whole process of photo generation $P(\mathcal{E}, \{x\})$ can be broken down into a series of 
sequential operations $O_k, k=1...N$, controlled by their own set of parameters $\{x_{O_k}\}$.
Each $O_k$ operates on its subject to yield a new photo:

\begin{equation}
P_k = O_k(\{x_k\}) \circ .... \circ O_1(\{x_1\}) \circ \mathcal{E}
\end{equation}

A well-defined and differentiable model of $\Phi$ will conclude our paper at this point. 
But in practice, it is next to impossible to obtain a dataset that defines $\Phi$.
Here are several reasons why:

\begin{itemize}
\item Curse of Dimensionality
\end{itemize}

By definition, $\Phi$ needs to incorporate all aesthetic aspects, such as saturation,
detail level, composition... 
To define $\Phi$ with examples, number of images needs to grow exponentially to
cover more aspects~\cite{acceptable}.

To make things worse, unlike traditional problems such as object recognition, what we need
are not only natural images, but also pro-level photos, which are much less in quantity.

\begin{itemize}
\item Meaningful Gradient
\end{itemize}

Even if $\Phi$ manages to approximate aesthetic quality using examples,
its gradient may not provide guidance on every aspect in aesthetics.
If a professional example was introduced mainly for its composition quality, it
offers little insight on HDR strength for similar photos.

\begin{itemize}
\item Before/After Pairs are Harder to Obtain
\end{itemize}

To put things to extremity, let's say all pro-level photos in this dataset have
a signature printed somewhere, while all lower-quality
photos don't.
$\Phi$ can be trained to high accuracy just by detecting signatures.
Obviously such $\Phi$ provides no guidance to our goal.
In reality it is similarly hard to force $\Phi$ to focus on aesthetic quality alone,
instead of other distribution imbalance between photos at different quality.

One useful trick is to provide photo-pairs before and after post-processing by photographers.
This way $\Phi$ is forced to only look at difference the photographer made.
However, it is even harder to collect such a dataset in large quantity.

\begin{itemize}
\item Hard to Optimize
\end{itemize}

It is difficult to optimize all aesthetic aspects in their joint high-dimensional space.

\subsection{Segmentation of $\Phi$} \label{sec:segmentation_phi}
In our paper, we resolve above issues by choosing operations $O_k$ to be approximately
``orthogonal" to each other. 
With that we segment $\Phi$ into semi-orthogonal components:
$
\Phi := \underset{k}{{\Sigma}} \Phi_k + \Phi_{res}
$
, where $\Phi_k$ only measures aesthetic changes that $O_k$ \emph{is capable of causing}.
This way applying $O_k$ has much less impact on all other $\Phi_j, j \ne k$, which makes
the optimization problem separable.

The objective function in Eq~\eqref{eq:objective} is approximated as:
\begin{equation}
\label{eq:new_objective}
\underset{k}{\Sigma} (\underset{\{x_{O_k}\}}{\arg\max} ( \Phi_i(P_i(\{x_{O_k}\})))) +  \Phi_{res}(P_N)
\end{equation}

Here each $\underset{\{x_{O_k}\}}{\arg\max} ( \Phi_k(P_k(\{x_{O_k}\})))$ becomes a lot more
tractable.
Intuitively, if $O_k$ is an image filter that changes overall saturation, 
$P_k(\{x_{O_k}\}) := O_k(\{x_{O_k}\}) \circ P_{k-1}$ applies saturation filter,
controlled by $\{x_{O_k}\}$, on a constant input photo $P_{k-1}$ from last step.
$\Phi_k$ is a metric that \emph{only} cares about how much saturation is aesthetically
pleasing for a photo.
The optimization seeks the right saturation amount that satisfies $\Phi_k$.
Such optimization happens in the dimension of $\{x_{O_k}\}$, which is much lower comparing to Eq~\eqref{eq:objective},
sometimes even in $1d$.

$\Phi_{res}$ captures all remaining aesthetic aspects missed by $\{\Phi_k\}$.
However, if $\{O_k\}$ exhausts our operations, whose aesthetic effects were already measured
by $\{\Phi_k\}$, there is literally nothing we can do to improve $\Phi_{res}$.

However, aesthetic aspects in $\Phi_{res}$ can still be captured to evaluate overall aesthetic quality.
We train a scorer  
$\Phi^{'} \sim \Phi :=\underset{k}{{\Sigma}} \Phi_k + \Phi_{res}$
using Inception v3 ~\cite{DBLP:journals/corr/SzegedyVISW15} to predict AVA ranking scores (see Section~\ref{sec:scores})
directly from images.
This scorer is later used to rank created photos.

\subsection{Selection of Operations $\{O_k\}$} \label{sec:ops_selection}

The intuition behind the requirement for $\{O_k\}$ to be ``orthogonal", is that we don't
want a later step $O_j$ to damage aesthetic quality $\Phi_i$ that was optimized by an earlier
step $O_i$.
But in practice, the pixel change caused by any image filter is hardly 
orthogonal to that of another.
We use this term loosely, and manually pick operations $\{O_k\}$ in following order:

\begin{itemize}
\item Composite an image from environment
\item Apply saturation filter
\item Apply HDR filter
\item Apply dramatic mask
\end{itemize}

Their effects are roughly independent to each other, focusing on composition,
saturation, detail level and low frequency brightness variation, respectively.
In section \ref{sec:saturation_hdr_experiment} we will show how this
semi-orthogonality helps each $\Phi_k$ to focus on its respective aesthetic aspect.

\subsection{Operation-Specific Aesthetic Metrics $\{\Phi_{k}\}$} \label{sec:ops_metrics}

Let's use a saturation filter $O_s$ as an example of image operations.
We want to train a metric $\Phi_{s}$, that focuses only on
aesthetic quality related to saturation, not on anything else.

If a dataset contains photos with labeled overall aesthetic quality,
it can not be directly used to train $\Phi_{s}$ because
contribution of saturation is mixed with all other aesthetic aspects.
Instead, we propose a method that only uses pro-level photos as positive examples for saturation training,
with the maximum $\Phi_{s}$ score assigned to them.
We then randomly perturb saturation level in these photos using $O_s$.
Its difference to the original photo serves as a penalty on $\Phi_{s}$ for
the perturbed photo.

A deep-learning model is then trained to predict $\Phi_{s}$ for these photos.
Since a high-score photo and its low-score counter-part only
differ by the perturbation, the model focuses only on what $O_s$ did, nothing else.
This makes $\Phi_{s}$ much easier to train than $\Phi$.
While gradient in $\Phi$ can be ill-defined, we can now find gradient for 
each $\Phi_{k}$ using its respective $O_k$.

Since negative examples are generated on the fly, this method removes the need for 
before/after photo pairs in dataset.
We can start with a same set of professional photos to train different aspects in aesthetics.

The algorithm for training $\Phi_{k}$ is given below.
\begin{algorithm}[]
\begin{algorithmic}[1]
  \Require Dataset of professional photos $\{M\}$
  \Require Image operation $O_k$:  $M_{out} = O_k(\{x\}) \circ M_{in}$
  \Require A metric measures similarity between two photos: $Sim(M_1, M_2) \in [0, 1]$. 1.0 means identical.
  \State $D=\{\}$, a set to hold (image, score) pairs.
  \For{Each image $M$ in $\{M\}$}
    \State Insert $(M, 1.0)$ in $D$
    \State \# Randomly sample parameters within a range:
    \For{Each dimension d in $\{x\}$}

      offset$_d$ = Random(xmin$_d$, xmax$_d$)
    \EndFor 
    \State $M^{'} = O_k(\{$offset$\}) \circ M $
    \State Insert $(M^{'}, Sim(M, M^{'}))$ in $D$
  \EndFor
  \State Train a model to predict score $\Phi_{k}$ from $D$
\end{algorithmic}
\end{algorithm}

When applying an operation on a photo, we optimize filter parameters to maximize
$\Phi_{k}$ on the input photo.
When $O_k$ has only one parameter, the optimization becomes a fast $1d$
search.
In this project, we optimize Saturation and HDR filters using this method.

\section{Special Operations} \label{sec:special_ops}

Two special operations, cropping and dramatic mask, deviate slightly from the algorithm
in section~\ref{sec:ops_metrics}.

\subsection{Image Composition} \label{sec:crop}

In the first operation $O_{crop}$, our robot finds the best compositions from the environment. 
In our project, the environment is represented by a spherical panorama.
We first do 6 camera projections to sample the panorama:
each projection is separated by 60 degree to cover all directions, with pitch angle 
looking slightly up at 10 degree, field of view 90 degree.
This way each project overlaps with its neighbors to increase chance for any composition to
be contained in at least one projection.

$\Phi_{crop}$ is trained to pick the best crop for each of these 6 images.
Its training is similarly to other $\Phi_k$, 
where a perturbed image $M^{'}$ is a random crop from $M$. 
$Sim(M, M^{'})$ is simply defined as $\frac{Area(M^{'})}{Area(M)}$.
Both $M$ and $M^{'}$ are resized to a square of fixed size before training.

Intuitively, the score says that in a professional photo $M$, the photographer chose current composition
over a zoomed-in version to favor a better composition.
Since cropping only removes content, $M^{'}$ never creates a case when a good composition is 
encompassed by unnecessary surroundings.
While techniques like image inpainting can potentially fake some surrounding to a photo, in practice
we noticed $\Phi_{crop}$ performs reasonably well without such cases.

However, cropping needs special treatment because the correlation between its operation $O_{crop}$ and other $\Phi_k$. 
Intuitively, if we change saturation of a photo, $\Phi_{crop}$ is
designed to remain indifferent.
But if we crop an image to a small patch, $\Phi_{s}$ may suffer a lot,
because image content is totally changed.

Thus we can not turn a blind eye to other aesthetic aspects during cropping.
We instead use 

\begin{equation}
\label{eq:crop_metric}
\Phi^{'}_{crop}(c) := c \times \Phi_{crop} + (1 - c) \times \Phi^{'}
\end{equation}
$\Phi^{'}$ is an approximate to $\Phi$ that judges overall aesthetic from Section~\ref{sec:segmentation_phi}. 
$c$ is a constant to weigh in the
importance of composition vs. overall aesthetic quality.
In Section ~\ref{sec:train_crop}, experiment shows that such a hybrid metric leads to better cropping options.

During optimization, a sliding window at different size and aspect ratio searches for crops with high
$\Phi^{'}_{crop}(c)$.

\subsection{Dramatic Mask} \label{sec:dramatic_mask}

A photographer often needs to stay at one spot for hours, waiting for the perfect
lighting.
When that doesn't work out, post-process may also add dramatic or even surreal lighting to a photo.
Vignetting is a commonly used filter that modulates brightness spatially with fixed
geometry, usually brighter near the center, darker at boundary.
Changing lighting based on image content is typically a manual job.

We want to learn a novel image operation, dramatic mask, that enhances dramatic lighting by
modulating brightness gradually in a photo. 
Good examples with such lighting must exist in $\{M\}$.
However, $\{M\}$ contains no additional label to indicate that.
We show that it is still possible to learn such a specific operation from unlabeled dataset.

Since we do not have an existing $O_{dramatic}$ to create $\{M^{'}\}$ as negative examples,
we use a list of existing image filters $\{F_j\}$ that are capable to change brightness in
various ways:

$M^{'} =$ Randomly pick from $\{ F_j \circ M\}$

The selection of $\{F_j\}$ is discussed in section \ref{sec:dramatic_mask_experiment}.
Here the assumption is $M$, being a professional photo, has a chance of containing dramatic lighting.
If the lighting in the photo is changed significantly by any $F_j$, it will likely lead to 
less ideal lighting.

Since $O_{dramatic}$ does not exist yet, we no longer separate the training of $\Phi_{dramatic}$
from the applying of $O_{dramatic}$ as in section \ref{sec:ops_metrics}.
Instead, they are trained jointly in a Generative-Adversarial Network (GAN).

GAN has demonstrated great capability in creating apparently novel contents from examples.
In most cases such creation is seeded by random numbers or controlling parameters.
Its outcome is typically a fixed-size raster in image space, usually with some noticeable
artifacts.
Generating an image at arbitrary size without artifacts is currently challenging.

Instead of generating pixels directly, we target to generate the ``editing" operation $O_{dramatic}$
using GAN, which is less susceptive to artifacts.
This operation is parameterized by a low-frequency $8\times8$ mask to
modulates brightness of input photo $M^{`}$:

$O_{dramatic} \circ M^{'} := M^{'} + mask \times ($ Brighten $ \circ M^{'} - M^{'})$

Here ``Brighten" is an image filter that makes input brighter.
During training mask is smoothly up-scaled to the size of $M^{`}$ using bilinear interpolation.

\begin{figure}[]
\begin{subfigure}{.48\textwidth}
\centering
\includegraphics[width=1\linewidth]{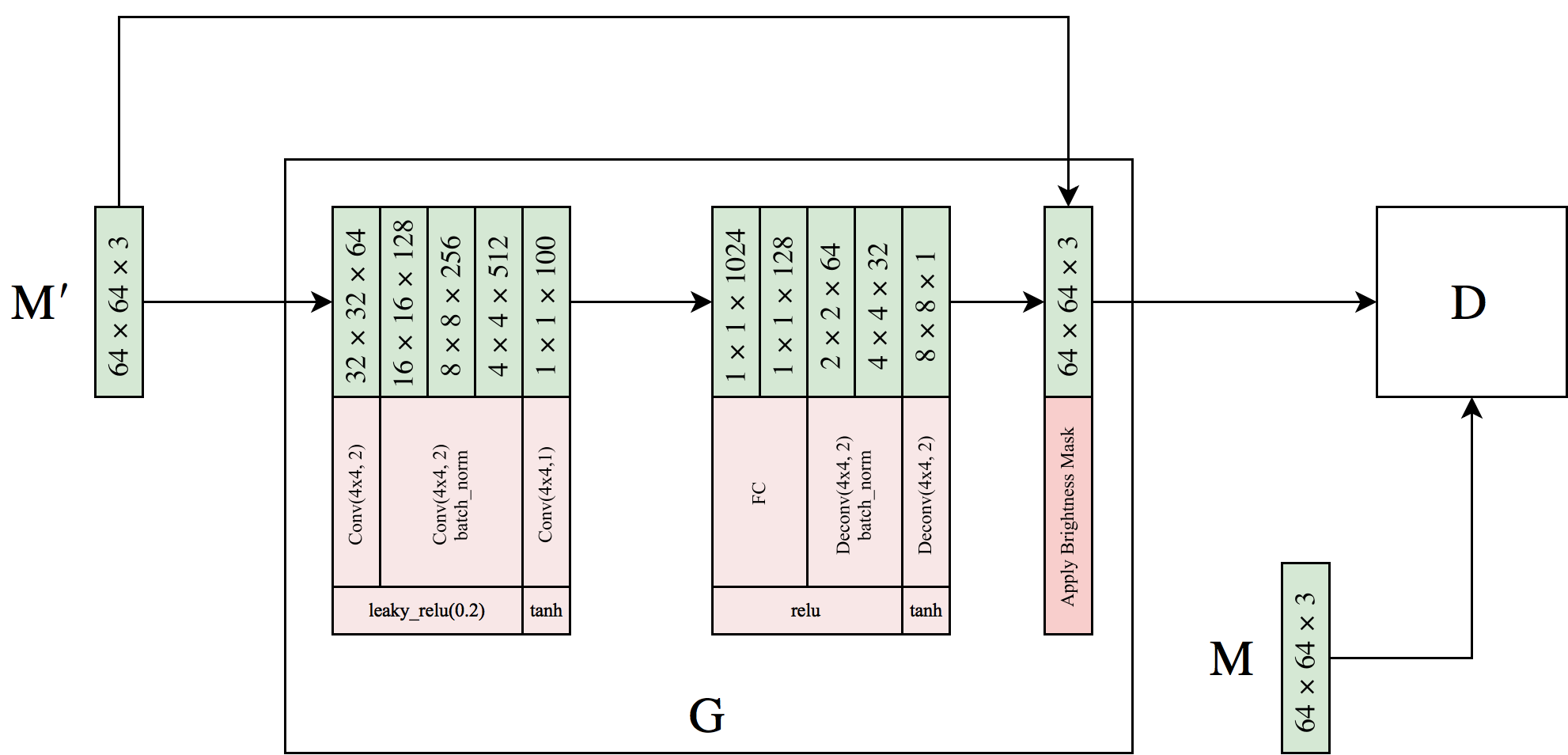}
\end{subfigure}
\caption{GAN structure for dramatic mask training. }
\label{fig:dcgan}
\end{figure}

We learn how to generate the mask in the generative model G.
The overall network architecture is DCGAN-based~\cite{DBLP:journals/corr/IsolaZZE16},
with G conditioned by our input image.
Its network architecture is depicted in Figure~\ref{fig:dcgan}.

The discriminative model D tries to distinguish images from $\{M\}$ and $\{O_{dramatic} \circ M^{'}\}$.
To encourage variations in G, we do not have a loss that compares pixel-difference between 
$O_{dramatic} \circ M^{'}$ and the photo $M^{`}$ was derived from.
Intuitively, there should exist many different ways to change the lighting in a scene
such that it becomes more dramatic.

Due to the competitive nature of GAN, it is very difficult for G to converge to a static state~\cite{NIPS2014_5423}.
Instead of waiting for an optimal mask, we use GAN to provide multiple candidates.
We train multiple models with different random initialization, collecting snapshots over time. 
All snapshots form a set of candidate models $\{\mathcal{M}\}$.
In the end, $\Phi^{'}$ is used to pick the best result generated from all members of $\{\mathcal{M}\}$.

Overall, the algorithm of how our robot operates is:

\begin{algorithm}[]
\begin{algorithmic}[1]
  \Require The robot ``travels" through environments $\{\mathcal{E}\}$ 
  \State Results $\{R\}=\{\}$
  \State Photos to process $\{P\}=\{\}$.
  \For{Each environment $\mathcal{E}$ in $\{\mathcal{E}\}$}
    \State Project $\mathcal{E}$ to 6 images $\{P_0\}$ in different directions 
    \For{Each projected image $P_0$ }
      \For{composition importance c in $\{0, 0.5, 1\}$}
    	\State Find best crops $\{P_1\}$ using $\Phi^{'}_{crop}(c)$
    	\State Insert top $k=3$ crops of $\{P_1\}$ into $\{P\}$
      \EndFor 
    \EndFor 
  \EndFor 
  \For{Each $P_1$ in $\{P\}$}
    \State Maximize $\Phi_{HDR}$: $P_2 = O_{HDR} \circ P_1$
    \State Maximize $\Phi_{saturation}$: $P_3 = O_{saturation} \circ P_2$
    \State Dramatic mask results $\{P_{\mathcal{M}}\}=\{\}$,
    \For{Each dramatic mask model $\mathcal{M}$ in $\{\mathcal{M}\}$}
      \State $P_{3_i}= O_{dramatic}(\mathcal{M}) \circ P_2$
      \State Insert $P_{3_i}$ to $\{P_{\mathcal{M}}\}$
    \EndFor 
    \State $P_3 = $ image in $\{P_{\mathcal{M}}\}$ with max $\Phi^{'}$.
    \State Insert $P_3$ in $\{R\}$
  \EndFor
  \State For crops in $\{R\}$ from same $P_0$, keep one with max $\Phi^{'}$
  \State Rank $\{R\}$ with $\Phi^{'}$.
\end{algorithmic}
\end{algorithm}

\section{Aesthetic Scale}\label{sec:scores}

In this paper, we target to generate photos towards professional quality
on a measurable aesthetic ``scale",
such that they can be comparable with all other photos in the world.
We work with professional photographers to empirically define 4 levels of aesthetic quality:

\begin{itemize}
\item 1: point-and-shoot photos without consideration.
\item 2: Good photos from the majority of population without art background. Nothing artistic stands out.
\item 3: Semi-pro. Great photos showing clear artistic aspects. The photographer is on the right track
 of becoming a professional.
\item 4: Pro-work.
\end{itemize}

Clearly each professional has his/her unique taste that needs calibration.
We use AVA dataset to bootstrap a consensus among them.
All AVA images are sorted by their average voting scores.
A single percentage ranking score between $[0, 1]$ is then assigned to each image.
For example, an image with ranking score $0.7$ is ranked at top $30\%$ among all AVA images.

We empirically divide this ranking score into 4 levels using thresholds $0.15, 0.7, 0.85$,
to roughly correspond to our aesthetic scale.
This correspondence is by no means accurate.
It only encourages a more even distribution of sampled AVA images across different qualities.
We sample images evenly from 4 levels, and mix them randomly.

Each professional is asked to score these images based on the description of our aesthetic scale.
After each round, we find average score for each photo as the consensus.
For individual score deviating a lot from the consensus, we send the photo
with consensus to the corresponding professional for calibration.
After multiple rounds we noticed a significant drop in score deviation, from 0.74 to 0.47.

At the end of this project, we asked professionals for their own descriptions of our aesthetic scale. 
Feedback is summarized in Table~\ref{table:scores_feedback}.

To map AVA ranking score $\Phi^{`}$ to aesthetic scale in range $[1, 4]$, we fit
$\bar{\Phi} = a \times \Phi^{`} + b$ to professionals' scores on AVA dataset.
$\bar{\Phi}$ is used in experiment to predict scores for each image.

\section{Experiments}\label{sec:experiments}
\subsection{Dataset}

For professional landscape photos $\{M\}$, we collected $\sim 15000$ thumbnails,
from high-rating landscape category on 500px.com.
Our trainings use thumbnails at a resolution up to $299 \times 299$ pixel.

To make overall style more consistent, we choose to target our goal
as ``colorful professional landscape".
We removed images from dataset that are black\&white or low in saturation, with
minimum average saturation per pixel set at 55\%.
Similar approach can be used to train towards other styles, such as ``black\&white
landscape".

AVA dataset ~\cite{6247954} was used to train $\Phi^{'}$, as well as for calibration 
of professionals' scores.

\subsection{Training of Cropping Filter}\label{sec:train_crop}

$M^{'}$ is randomly cropped from $M$ in two batches.
The first batch concentrates on variations close to original image, so we have
more examples to learn when composition is close to optimal.
The cropped width is between $(90\%, 100\%)$ of original image.
Aspect ratio is randomly selected between $(0.5, 2)$, with the area contained within image space.
The second batch deviates more from optimal composition, with width range $(50\%, 90\%)$.
Both batches are equal in number.
Score is defined as area ratio, as described in section \ref{sec:crop}.
The training network is Inception v3~\cite{DBLP:journals/corr/SzegedyVISW15}, 
which is used to predicts score from input image.

For each projected photo from panorama, 
we pick top 3 candidates at each composition weight $c \in \{0, 0.5, 1\}$.
All these candidates move forward in pipeline independently.
At the end of the pipeline, the candidate with highest $\Phi^{'}$ is selected to represent
that photo.

To further compare effects of composition weights, we conducted a separate experiment,
where 4 professionals are presented with 3 cropped version of a same photo, using the top candidate
from each composition weight $c \in \{0, 0.5, 1\}$.
They are asked to pick the one with best composition, or select ``none".
For all 100 input images, after excluding $10.4\%$ ratings of ``none", $22\%$ images received 
a unanimous voting on one cropping candidate. 
The distribution of $c \in \{0, 0.5, 1\}$ for winning candidates are $9.2\%$, $47.4\%$ and $43.4\%$,
respectively.
$80\%$ images has up to two winning candidates.
The distribution of winners are $13.2\%$, $41.4\%$ and $45.4\%$, respectively.
This shows that with our hybrid approach, we can produce better cropping candidates than using
a single metric.

\subsection{Training of Saturation and HDR Filters}\label{sec:saturation_hdr_experiment} 

In our approach the choice of $\{O_k\}$ is flexible. 
For saturation filter we used an implementation similar to the saturation adjustment of
``Tune Image" option in Snapseed.
During training $M^{'}$ is obtained by setting saturation parameter randomly between
$(0\%, 80\%)$ of filter's range, where $0\%$ turns a photo to black \& white.
Per-pixel-channel color difference $\delta$ is used to derive saturation score, with maximum difference
capped at $6\%$ for a score $0$:

  \[
    score := Sim(M, M^{'}) :=\left\{
                \begin{array}{ll}
                  1-\frac{\delta}{6\%}, \delta <= 6\%\\
                  0, otherwise
                \end{array}
              \right.
  \]

For each $M$, 6 variations of $M^{'}$ were generated for training, together with $M$.

HDR filter was trained in a same way, using an implementation similar to ``HDR Scape" in
Snapseed.
``Filter Strength" is the only parameter we modify.

One difference here is ``Filter Strength" only goes positively.
While we can reduce saturation of an image, there is no option to add ``negative" HDR effect
, which is the direction we care more.
Intuitively, we assume HDR effects already exists in some $M$. 
We want to generate $M^{'}$ that contains ``less" HDR, so we can learn how an HDR filter can
help it to look more like $M$.
We used a simple trick to mimic ``negative" HDR effect by a per-pixel operation:
\begin{equation}
\label{eq:negate}
F(-strength) \circ M = 2 \times M - F(strength) \circ M
\end{equation}

This way we expanded the range of ``Filter Strength" from $(0, max)$ to $(-max, max)$.
We generate $\{M^{'}\}$ in two batches. 
For each $M$, 
we sample 6 variations of not-enough-HDR examples $M_1^{'}$,
with ``Filter Strength" in range $(-max, -0.5max)$.
We also sample 3 variations of too-much-HDR examples $M_2^{'}$,
with ``Filter Strength" in range $(0.5max, max)$.
Color difference $\delta$ is capped at 20\%.

Training of both aesthetic metrics are same as that of cropping metric.

During optimization for each photo, a quick $1d$ search on parameter is applied for each filter.
Saturation parameter is tried from 40\% to 90\%, with step 10\%.
For HDR, ``Filter Strength" parameter is tried from 0\% to 70\% of maximum range, with step 10\%.
The parameter of each filter with highest score is committed to apply on the photo.

\begin{figure*}[t]
\begin{subfigure}{.16\textwidth}
\centering
\includegraphics[width=1\linewidth]{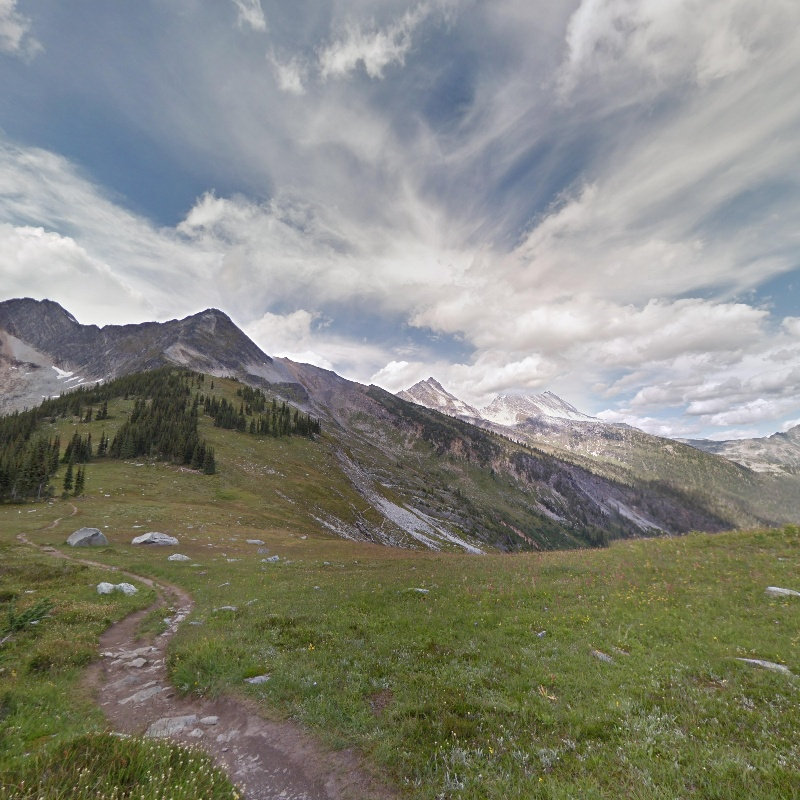}
\caption{Input image}
\label{fig:image_10}
\end{subfigure}
\begin{subfigure}{.27\textwidth}
\centering
\includegraphics[width=1\linewidth]{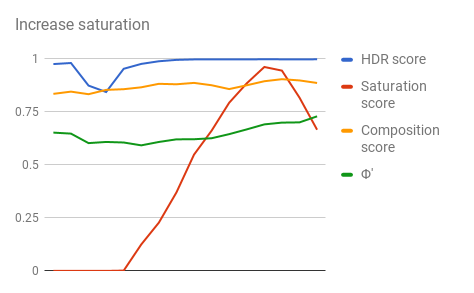}
\caption{Increase saturation to max}
\label{fig:s_change}
\end{subfigure}
\begin{subfigure}{.27\textwidth}
\centering
\includegraphics[width=1\linewidth]{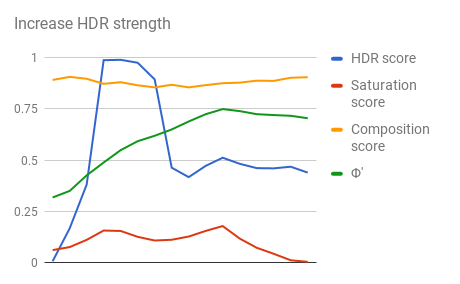}
\caption{Increase HDR strength to max}
\label{fig:hdr_change}
\end{subfigure}
\begin{subfigure}{.27\textwidth}
\centering
\includegraphics[width=1\linewidth]{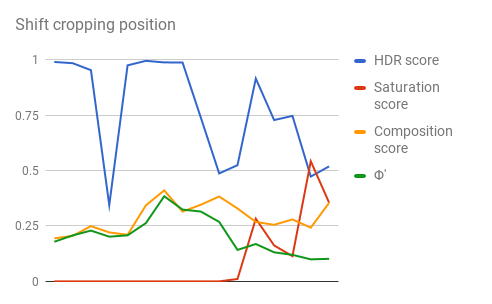}
\caption{Crop image with a sliding window from top to bottom}
\label{fig:cropping_change}
\end{subfigure}
\caption{Scores change when image (a) is modified by different operations. }
\label{fig:scores_change}
\end{figure*}

In Figure~\ref{fig:scores_change}, an example of changes in $\{\Phi_k\}$ from different image operations 
are visualized.
Saturation of input image is increased from $0$ to maximum in Figure~\ref{fig:s_change}.
Saturation score shows a distinctive peak, while scores for HDR and composition remain more flat.

Note that overall aesthetic score $\Phi^{'}$ shows continuous increasement, well into saturation range that makes the image over-saturated.
This is expected, because the training of $\Phi^{'}$ on AVA dataset only involves natural images.
Since a large portion of higher-score images are more-saturated than typical user photos directly
from a cell phone, $\Phi^{'}$ grows higher with increasing saturation.
However, since AVA does not intentionally implant over-saturated images as negative examples, 
$\Phi^{'}$ does not penalize over-saturated images.
This shows that a naively-trained general aesthetic metric may not be suitable for optimizing different aesthetic aspects.

Similarly, in Figure~\ref{fig:hdr_change} HDR score shows a clear peak when HDR strength increases.
Once again $\Phi^{'}$ remains high incorrectly even when HDR strength is too strong.

Figure~\ref{fig:cropping_change} shows the special case of image composition.
Here a sliding window of half image width moves vertically across the image at center,
with aspect ratio $1.8$.
Composition score contains a peak near image center.
However, both saturation and HDR scores vary a lot.
This is also expected, because image statistics regarding saturation and detail level both change
as the sliding window moves.
In other words, cropping operation is less ``orthogonal" to other image operations, which leads to our
hybrid approach on composition score.

\begin{figure}[t]
\begin{subfigure}{.16\textwidth}
\centering
\includegraphics[width=1\linewidth]{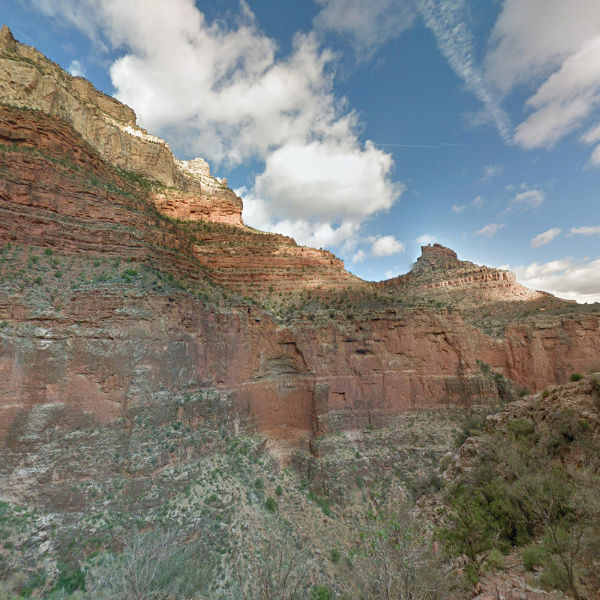}
\caption{Input image}
\label{fig:scores_coupled}
\end{subfigure}
\begin{subfigure}{.27\textwidth}
\centering
\includegraphics[width=1\linewidth]{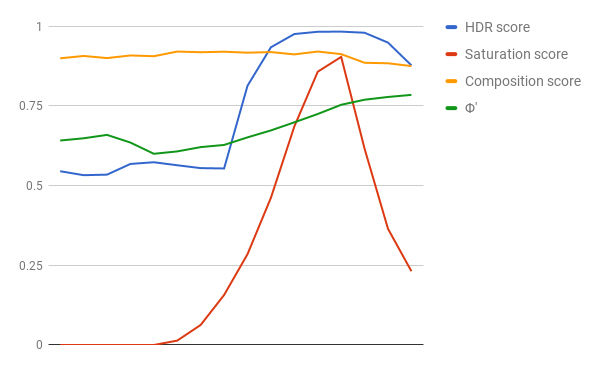}
\caption{Increase saturation to max}
\label{fig:scores_coupled}
\end{subfigure}
\caption{Saturation and HDR scores coupled when increasing saturation for image (a). }
\end{figure}

For some photo $\{\Phi_k\}$ may become less orthogonal.
For example, when pixels over-saturate, details are also lost, which may impact HDR score that measures
detail level.
Figure~\ref{fig:scores_coupled} shows such an example, where as saturation increases, HDR score starts 
to change too. 
In such a case, two linear search in saturation and HDR strength separately may not
yield optimal solution.
Global methods like gradient-descent can find better solution, at the cost of more expensive search.
In this paper we use separated linear-search to generate all results.

\subsection{Training of Dramatic Mask}\label{sec:dramatic_mask_experiment} 

We use following image filters with equal chance to generate $\{F_j\}$. 
(Note that they can be replaced by other reasonable alternatives.)
\begin{itemize}
\item Snapseed ``Tune Image", brightness parameter randomly from $(10\%, 45\%)$ and $(55\%, 90\%)$
\item Snapseed ``Tune Image", contrast parameter randomly from $(10\%, 45\%)$ and $(55\%, 90\%)$
\item Snapseed ``HDR Scape", parameter randomly from $(10\%, 100\%)$ and $(-100\%, -10\%)$. Negative effect is simulated using Eq~\eqref{eq:negate}
\item Snapseed ``Vignette", with ``Outer Brightness" in $(0\%, 35\%)$ to darken boundary,
and $(60\%, 70\%)$ to brighten boundary.
\item A curve-editing filter, where 6 control points evenly 
spanning brightness range, move within $(-15\%, 15\%)$ of total range.
\item A flatten-brightness filter with strength $(10\%, 100\%)$, and strength $(-100\%, -10\%)$
using Eq~\eqref{eq:negate}
\end{itemize}

The flatten-brightness filter moves each pixel's brightness $b$ towards the Gaussian smooth average 
of its neighborhood $\overline{b}$ with strength $s \in (0, 1)$:
$b_{new} = \overline{b} \times s + b \times (1 - s)$.
Gaussian radius is $0.05 \times ($width $+$ height$)$.

We trained 37 models with different random initialization for about 2,000,000 steps each.
For each model, we collect snapshot of trained parameters roughly every 100000 steps.
We then apply all these models on a test dataset, and pick best mask for each image using $\Phi^{'}$.
We rank models by number of final results they contributed, and keep top 50.
After manual examination, 20 of these models, especially earlier in training, generate
visible artifacts to different degrees.
After removing them, we eventually keep top 30 models as $\{\mathcal{M}\}$.

\begin{figure}[t]
\begin{subfigure}{.15\textwidth}
\centering
\includegraphics[width=1\linewidth]{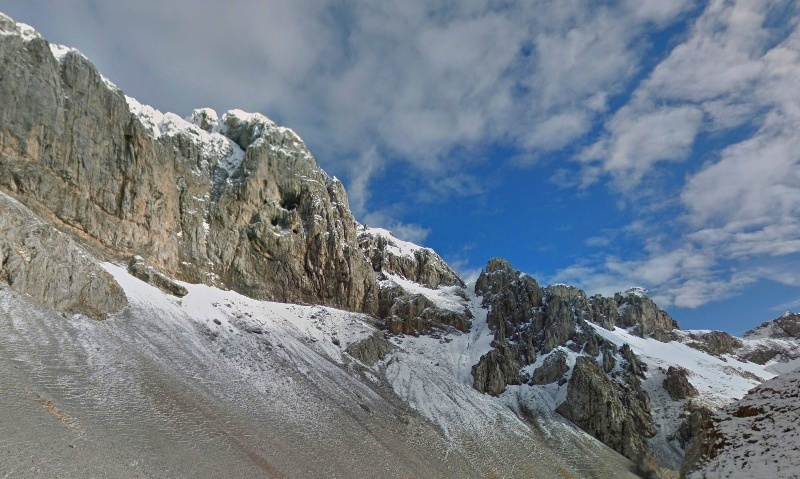}
\caption*{}
\end{subfigure}
\begin{subfigure}{.15\textwidth}
\centering
\includegraphics[width=1\linewidth]{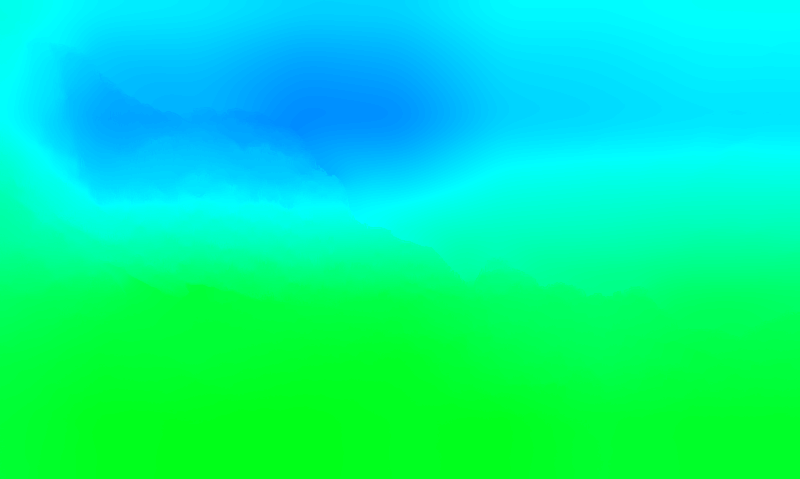}
\caption*{}
\end{subfigure}
\begin{subfigure}{.15\textwidth}
\centering
\includegraphics[width=1\linewidth]{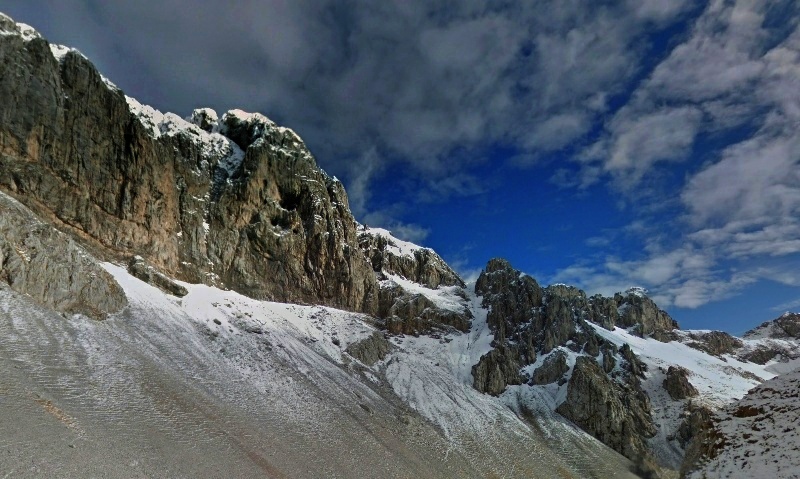}
\caption*{}
\end{subfigure} \\
\begin{subfigure}{.15\textwidth}
\centering
\includegraphics[width=1\linewidth]{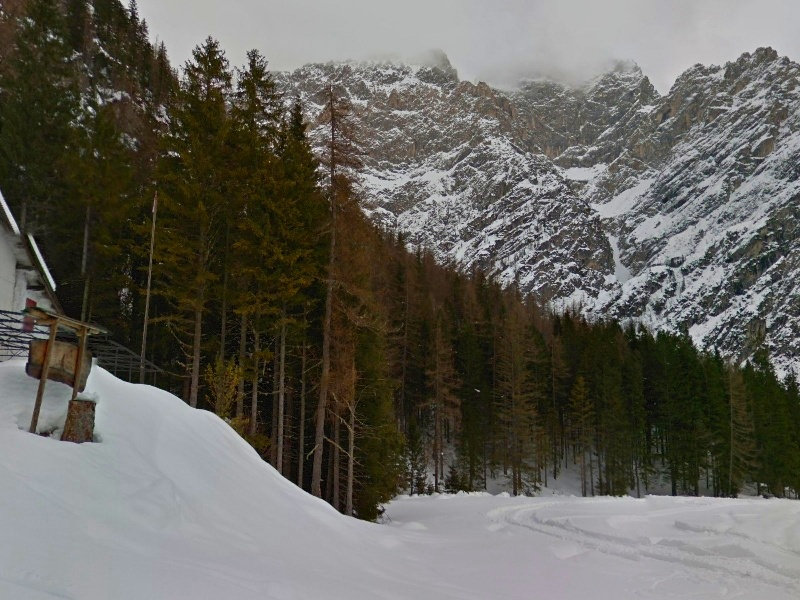}
\caption*{}
\end{subfigure}
\begin{subfigure}{.15\textwidth}
\centering
\includegraphics[width=1\linewidth]{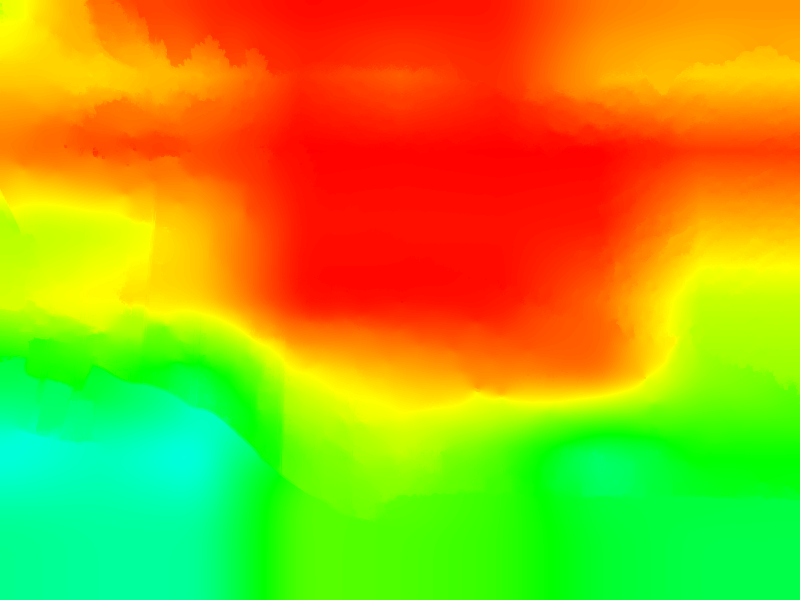}
\caption*{}
\end{subfigure}
\begin{subfigure}{.15\textwidth}
\centering
\includegraphics[width=1\linewidth]{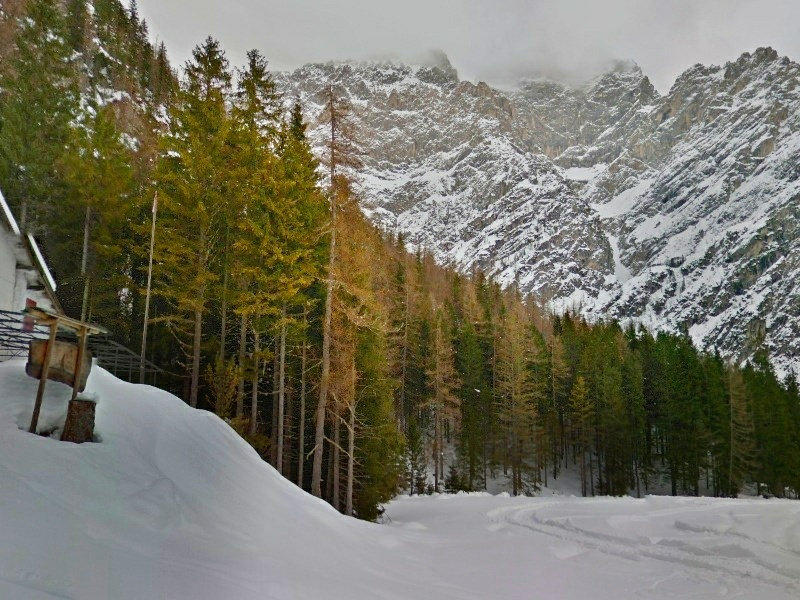}
\caption*{}
\end{subfigure} \\
\begin{subfigure}{.15\textwidth}
\centering
\includegraphics[width=1\linewidth]{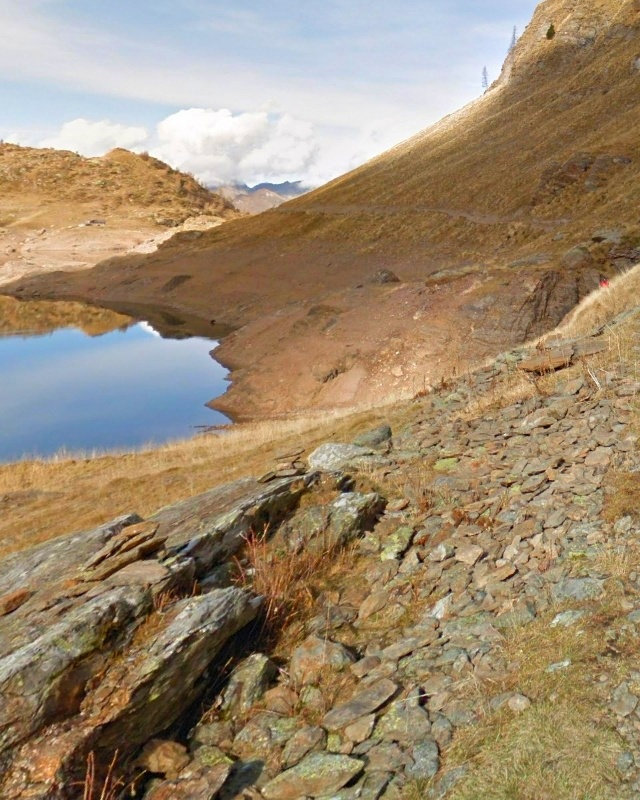}
\caption*{}
\end{subfigure}
\begin{subfigure}{.15\textwidth}
\centering
\includegraphics[width=1\linewidth]{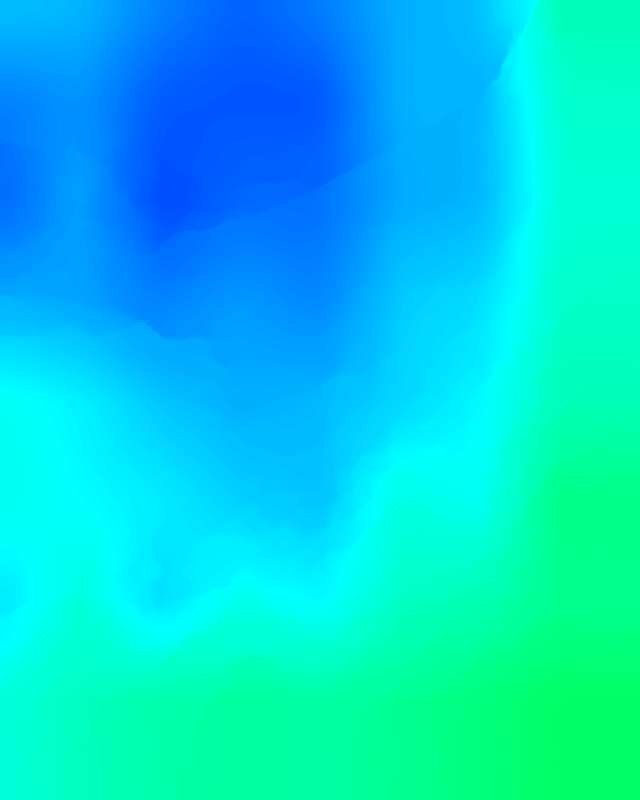}
\caption*{}
\end{subfigure}
\begin{subfigure}{.15\textwidth}
\centering
\includegraphics[width=1\linewidth]{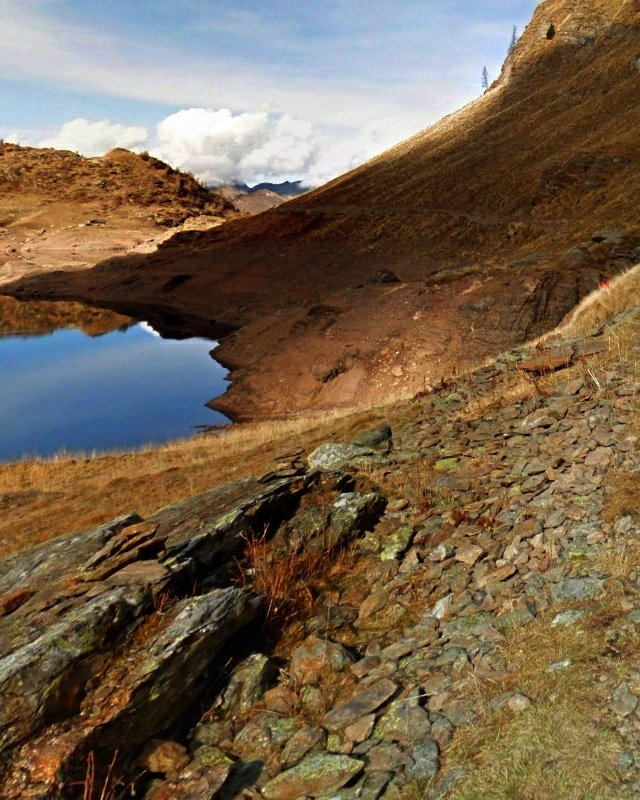}
\caption*{}
\end{subfigure}
\caption{Images before and after applying dramatic mask. }
\label{fig:dramatic_mask}
\end{figure}

To better align $8 \times 8$ mask with photo content at full resolution, we apply a joint bilateral 
upsampling~\cite{KCLU07} on the mask against the photo before generating the final result.
The effects of dramatic mask varies by images.
Sometimes it mimics traditional Vignetting. 
But in more general cases the brightness modulation is based on image content.
Figure~\ref{fig:dramatic_mask} shows several examples.

\begin{figure}[t]
\begin{subfigure}{.48\textwidth}
\centering
\includegraphics[width=1\linewidth]{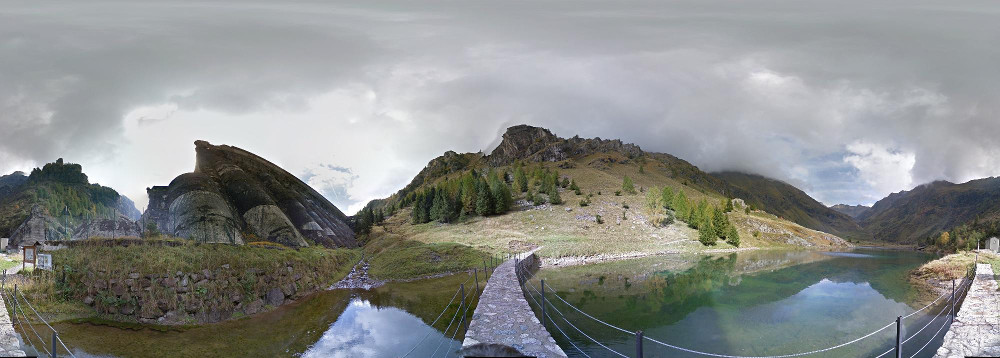}
\caption{}
\end{subfigure} \\
\begin{subfigure}{.156\textwidth}
\centering
\includegraphics[width=1\linewidth]{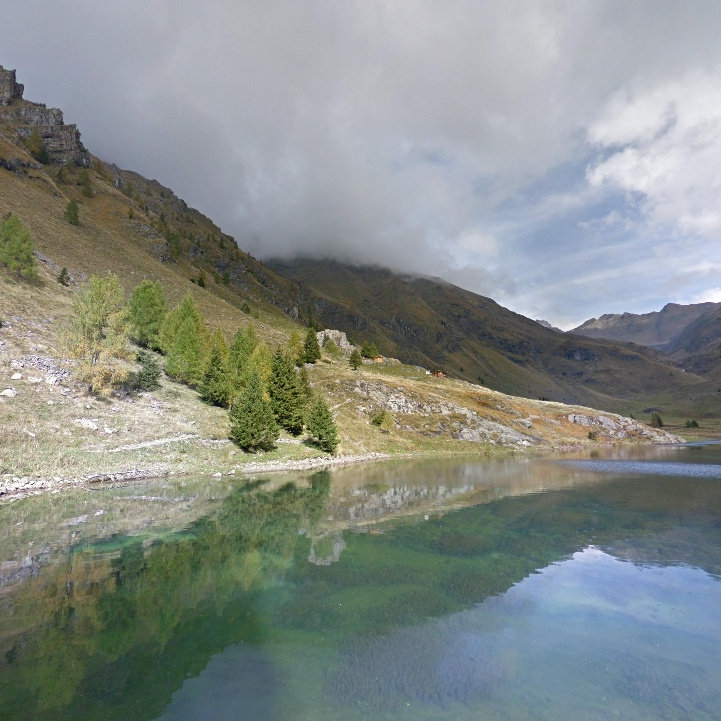}
\caption{}
\end{subfigure}
\begin{subfigure}{.156\textwidth}
\centering
\includegraphics[width=1\linewidth]{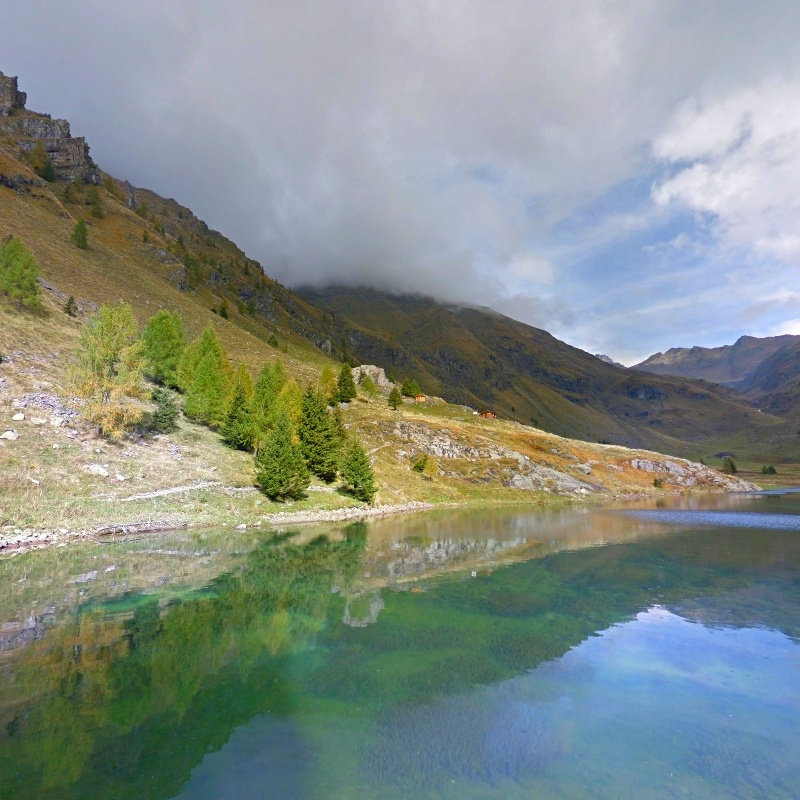}
\caption{}
\end{subfigure}
\begin{subfigure}{.156\textwidth}
\centering
\includegraphics[width=1\linewidth]{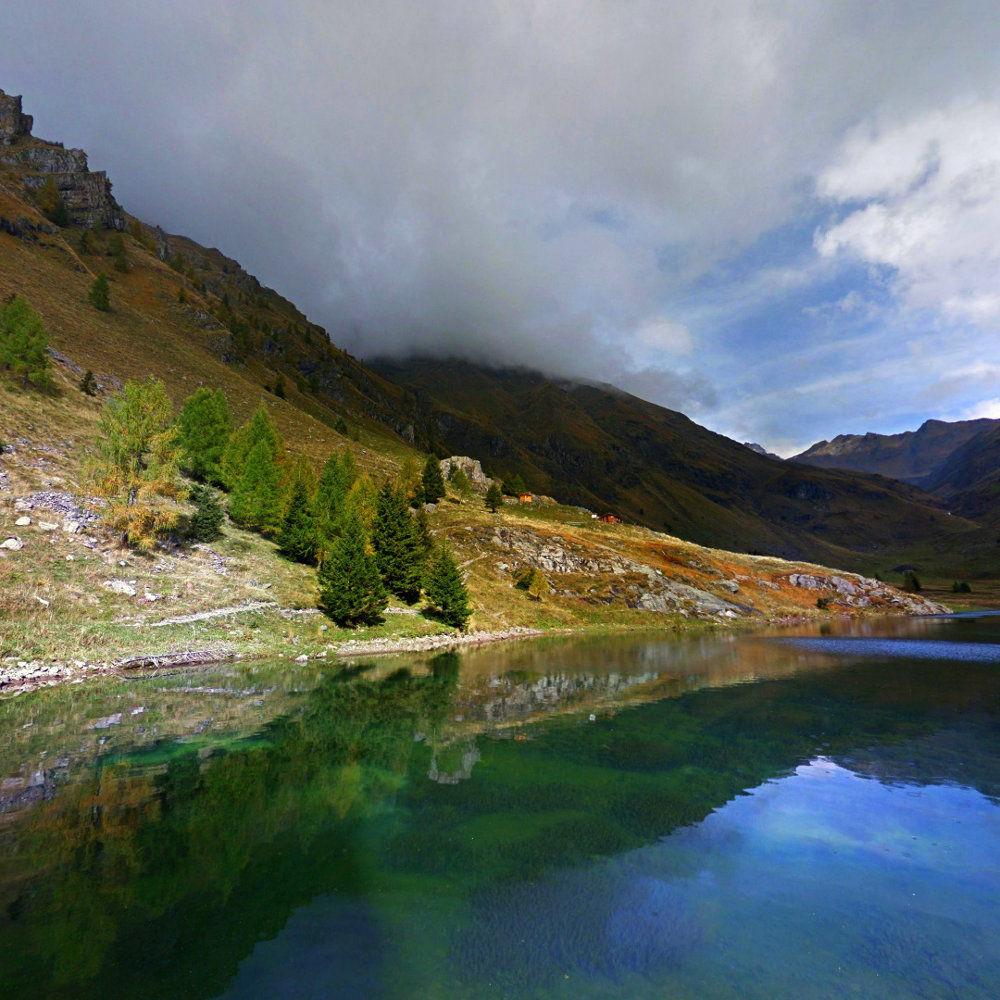}
\caption{}
\end{subfigure} \\
\caption{A panorama (a) is cropped into (b), with saturation and HDR strength enhanced in (c),
and with dramatic mask applied in (d). }
\label{fig:e2e}
\end{figure}

An end-to-end example is shown in Figure~\ref{fig:e2e}, where a panorama in (a) is cropped into (b),
with saturation and HDR strength enhanced in (c), and finally with dramatic mask applied in (d).

\subsection{Photo Creation}\label{sec:photo_creation} 

The environment for our robot is simulated by a collection of panorama images
from Google Street View.
Most trails we chose were collected on foot, instead of by vehicles.
Locations we picked including Banff, Jasper and Glacier national parks in Canada,
Grand Canyon and Yellowstone National Parks in US,
and foot trails in Alps, Dolomites and Big Sur.
Panoramas were sampled sparsely along each trail to reduce data amount and redundancy.
Neighboring panoramas are typically separated by tens of meters or more.
Altogether $\sim 40000$ panoramas were used.
From them $\sim 31000$ photos were created with predicted scores $\bar \Phi >= 2.5$.

Street View imagery we used contains some artifacts, which has an impact on quality
of our results.
Over-exposure happens quite often in brighter area, such as on cloud, which washes out details.
Misalignment and blurry parts can also be noticed in our results.

During evaluation, we manually removed results with
severe misalignment/blur, Street View equipments, black area after panorama stitching, 
large portion of highway and large human figures.

\subsection{Evaluation}\label{sec:evaluation} 

We work with 6 professionals for evaluation.
We select them with a minimum requirement of a bachelor's degree in Photography and
2+ years experience as a professional photographer.
To keep the evaluation as objective as possible, they were not informed of our 
image creation attempt.

We conducted 6 rounds of calibration as described in Section \ref{sec:scores},
using $\sim 2200$ AVA images in total.
The score deviation per image dropped from initial $0.74$ to $0.47$.

We randomly sampled $400+$ photos from our creation, with predicted scores
$\bar{\Phi}$ uniformly distributed between $(2.5, 3.0)$. 
(Most photos received a prediction score $< 3.0$ after linear fitting of $\bar \Phi$.)
They are randomly mixed with $800$ photos from AVA and other sources,
sampled across different quality.

\begin{figure}[]
\begin{subfigure}{.23\textwidth}
\centering
\includegraphics[width=1\linewidth]{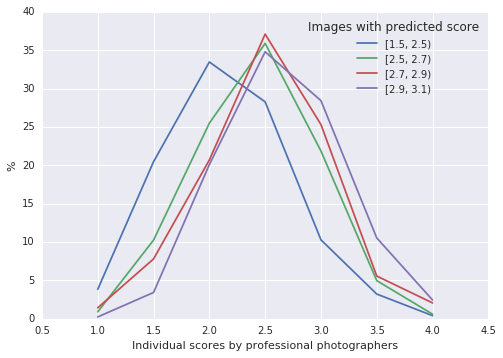}
\caption{}
\end{subfigure}
\begin{subfigure}{.23\textwidth}
\centering
\includegraphics[width=1\linewidth]{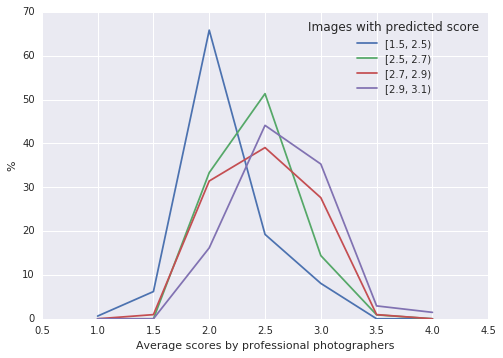}
\caption{}
\end{subfigure} \\
\begin{subfigure}{.23\textwidth}
\centering
\includegraphics[width=1\linewidth]{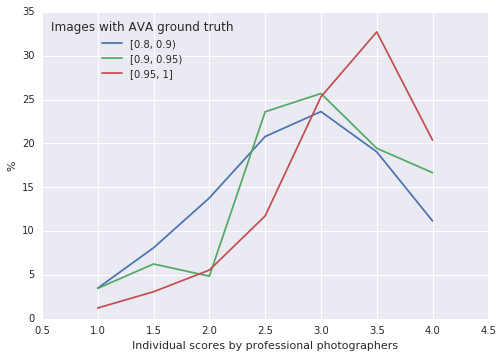}
\caption{}
\end{subfigure}
\begin{subfigure}{.23\textwidth}
\centering
\includegraphics[width=1\linewidth]{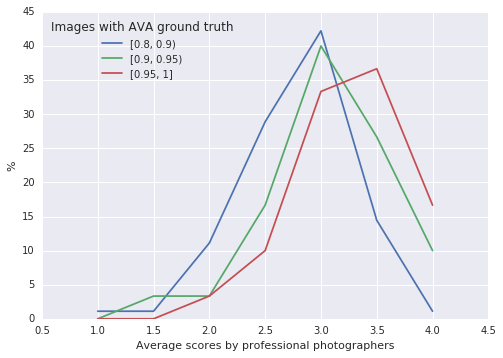}
\caption{}
\end{subfigure}
\caption{Individual (a) and per-image average (b) scores distribution for images at different predicted scores. As a comparison, score distribution for top-ranking AVA photos are shown in (c) and (d),
which are arguably a reasonable collection of actual professional work.}
\label{fig:pro_scores}
\end{figure}

Figure~\ref{fig:pro_scores} shows how scores distribute for photos at different predicted scores.
A higher predicted score leads to higher scores from professionals, which shows $\bar{\Phi}$ is correlated
to professional's taste.
More accurate prediction of professionals' scores is still to be achieved.
Average score per image is the average of individual scores from all 6 professionals for each image.

For $173$ evaluated photos with a high predicted score $>= 2.9$, $41.4\%$ of individual scores received from professionals
are at or above semi-pro level ($>= 3.0$), with $13\%$ of them $ >= 3.5$.

As a comparison, score distribution for sampled high-ranking AVA photos are also shown in Figure~\ref{fig:pro_scores} (c) and (d).
We can see even for top $5\%$ photos ($[0.95, 1]$ in graph) in AVA dataset, 
arguably all at professional quality, only $45\%$ scores are at or above $3.5$.
This may serve as an estimation of the ``ceiling" for such an algorithm.

\begin{figure}[t]
\begin{subfigure}{.14\textwidth}
\centering
\includegraphics[width=1\linewidth]{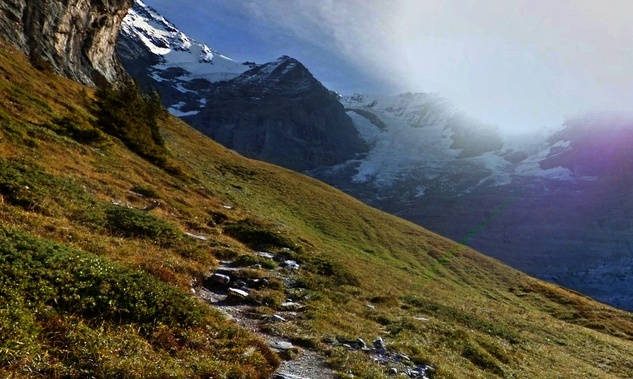}
\caption*{1.9}
\end{subfigure}
\begin{subfigure}{.14\textwidth}
\centering
\includegraphics[width=1\linewidth]{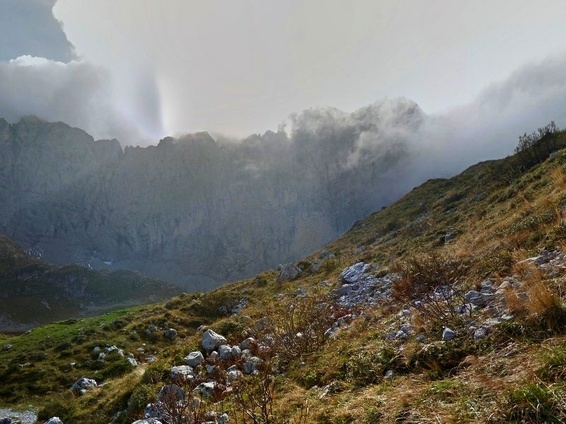}
\caption*{1.8}
\end{subfigure}
\begin{subfigure}{.14\textwidth}
\centering
\includegraphics[width=1\linewidth]{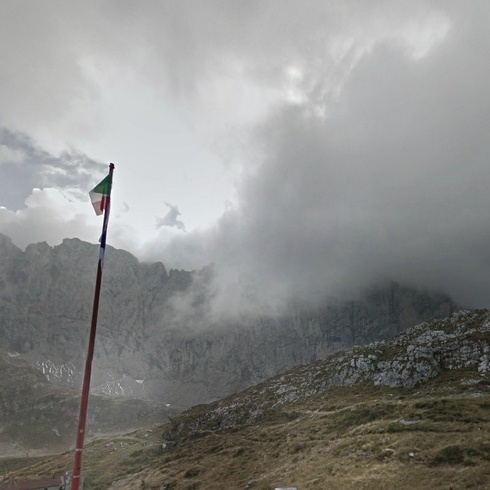}
\caption*{1.9}
\end{subfigure} \\
\begin{subfigure}{.14\textwidth}
\centering
\includegraphics[width=1\linewidth]{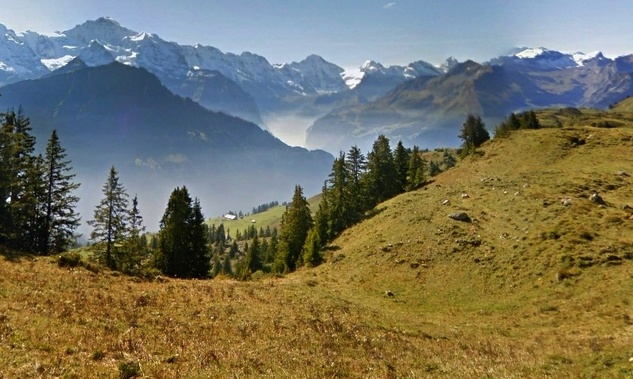}
\caption*{2.4}
\end{subfigure} 
\begin{subfigure}{.14\textwidth}
\centering
\includegraphics[width=1\linewidth]{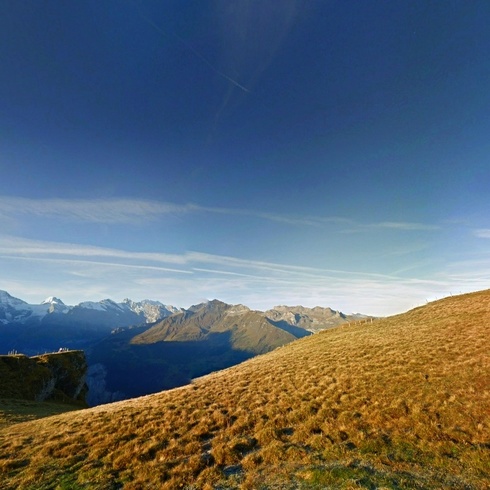}
\caption*{2.5}
\end{subfigure}
\begin{subfigure}{.14\textwidth}
\centering
\includegraphics[width=1\linewidth]{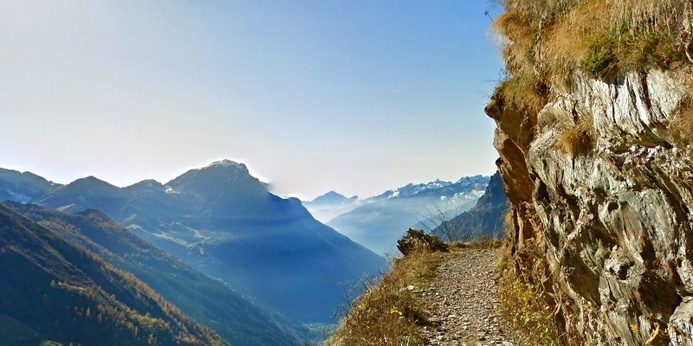}
\caption*{2.4}
\end{subfigure} \\
\begin{subfigure}{.14\textwidth}
\centering
\includegraphics[width=1\linewidth]{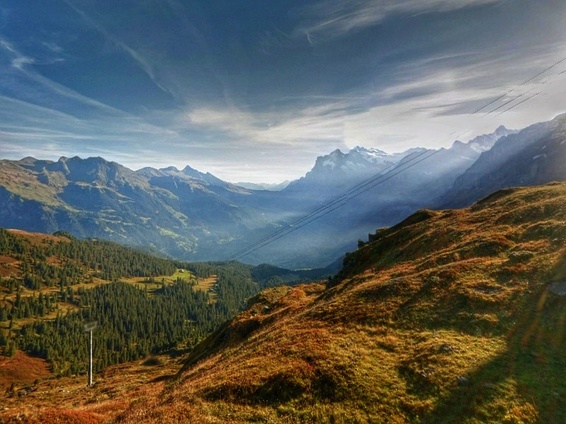}
\caption*{3.3}
\end{subfigure}
\begin{subfigure}{.14\textwidth}
\centering
\includegraphics[width=1\linewidth]{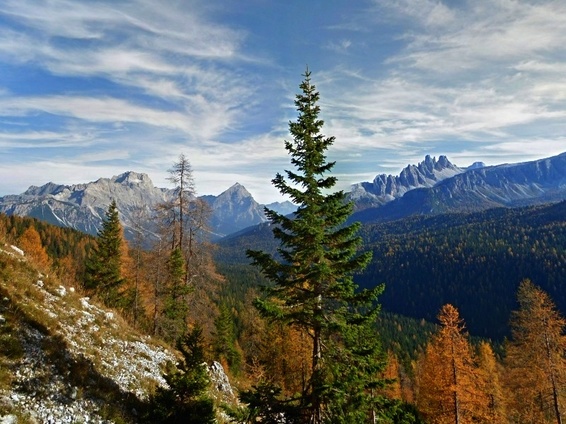}
\caption*{3.0}
\end{subfigure}
\begin{subfigure}{.14\textwidth}
\centering
\includegraphics[width=1\linewidth]{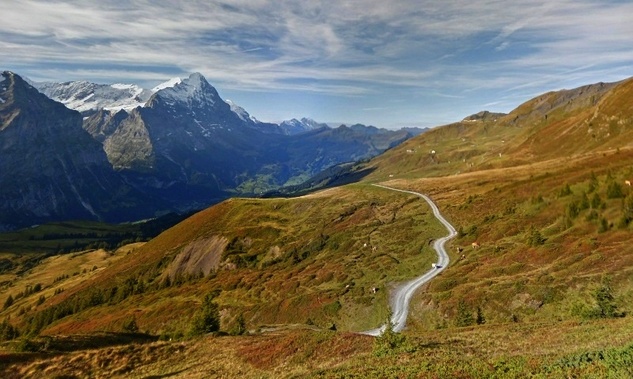}
\caption*{3.0}
\end{subfigure}\\
\caption{Example photos with predicted scores $> 2.7$. Average professional scores are displayed below.}
\label{fig:score_examples}
\end{figure}

Some examples of created photos with prediction scores $> 2.7$ are displayed in Figure~\ref{fig:score_examples}. 
To demonstrate their quality variation according to the professionals, 
we select photos with average professional scores $< 2.0$ in first row, 
$\sim 2.5$ in second row, and $> 3.0$ in last row.

\begin{figure*}[t]
\begin{subfigure}{.33\textwidth}
\centering
\includegraphics[width=1\linewidth]{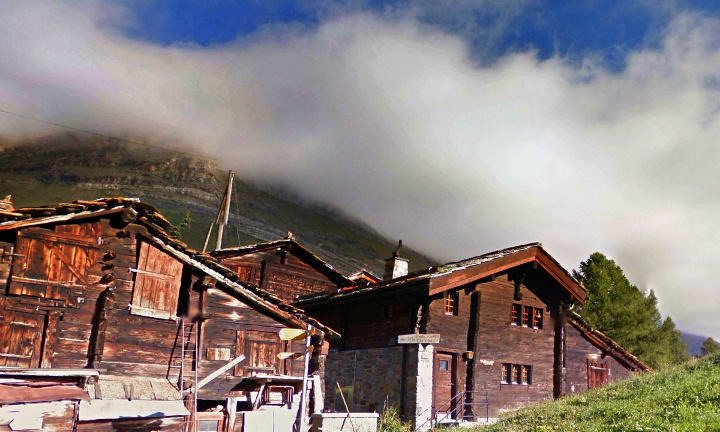}
\caption*{Predict: 2.4, Pro average: 2.8}
\end{subfigure}
\begin{subfigure}{.555\textwidth}
\centering
\includegraphics[width=1\linewidth]{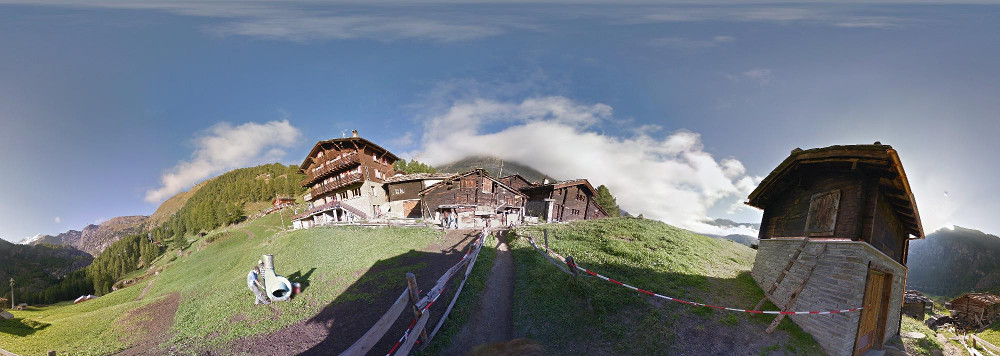}
\caption*{}
\end{subfigure} \\
\begin{subfigure}{.33\textwidth}
\centering
\includegraphics[width=1\linewidth]{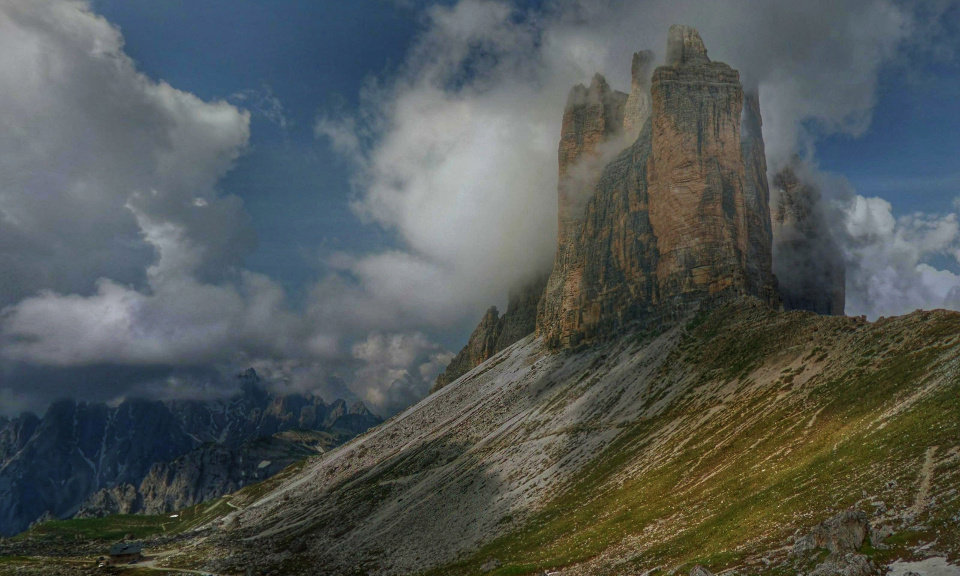}
\caption*{Predict: 2.7, Pro average: 3.0}
\end{subfigure} 
\begin{subfigure}{.555\textwidth}
\centering
\includegraphics[width=1\linewidth]{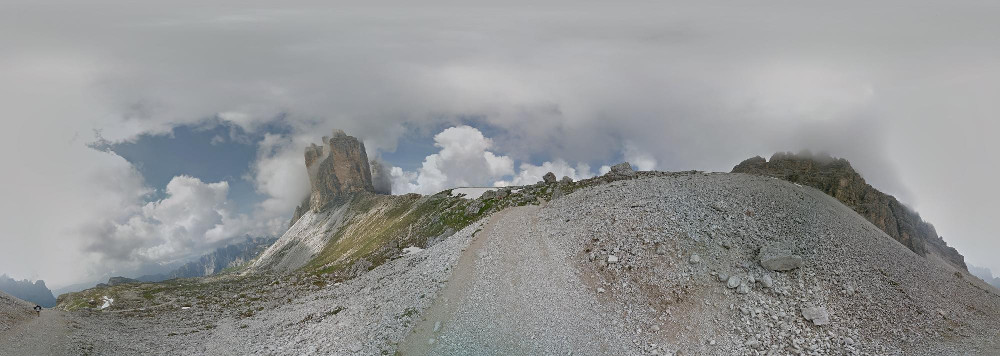}
\caption*{}
\end{subfigure} \\
\begin{subfigure}{.25\textwidth}
\centering
\includegraphics[width=1\linewidth]{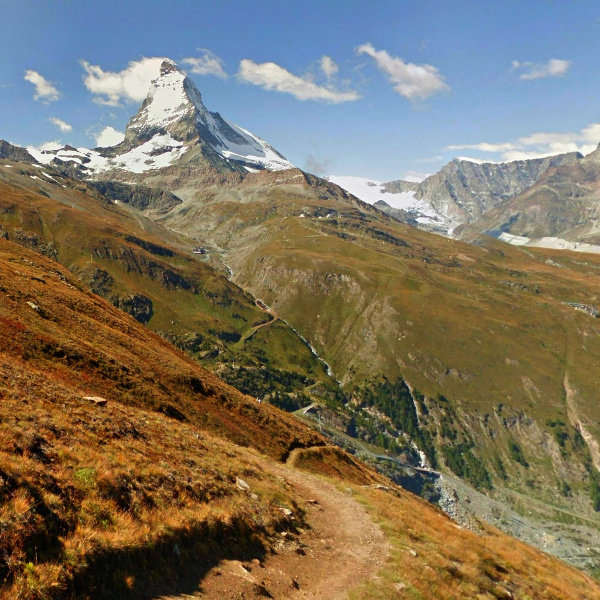}
\caption*{Predict: 2.6, Pro average: 3.2}
\end{subfigure}
\begin{subfigure}{.65\textwidth}
\centering
\includegraphics[width=1\linewidth]{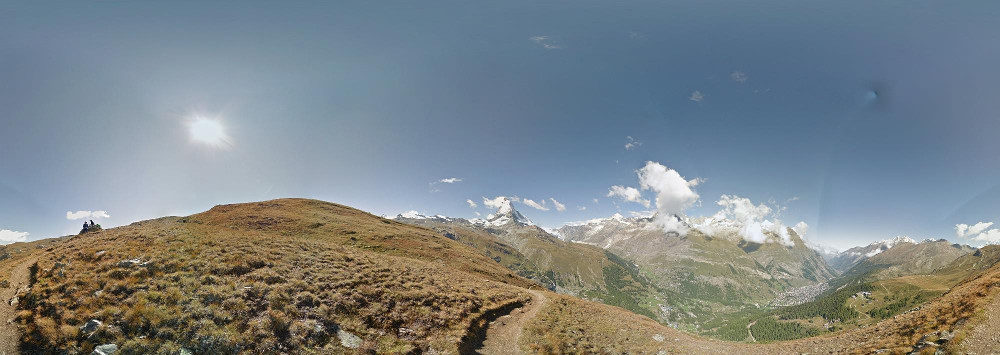}
\caption*{}
\end{subfigure} \\
\begin{subfigure}{.25\textwidth}
\centering
\includegraphics[width=1\linewidth]{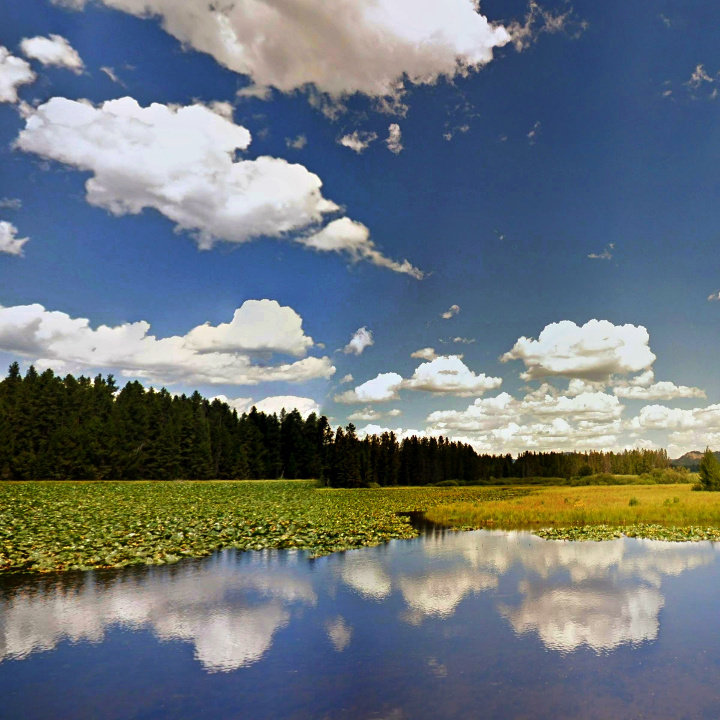}
\caption*{Predict: 2.8, Pro average: 2.5}
\end{subfigure}
\begin{subfigure}{.65\textwidth}
\centering
\includegraphics[width=1\linewidth]{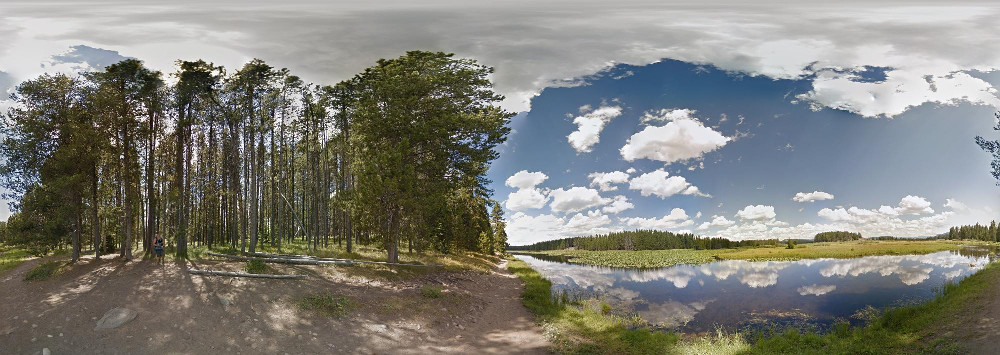}
\caption*{}
\end{subfigure} \\
\begin{subfigure}{.25\textwidth}
\centering
\includegraphics[width=1\linewidth]{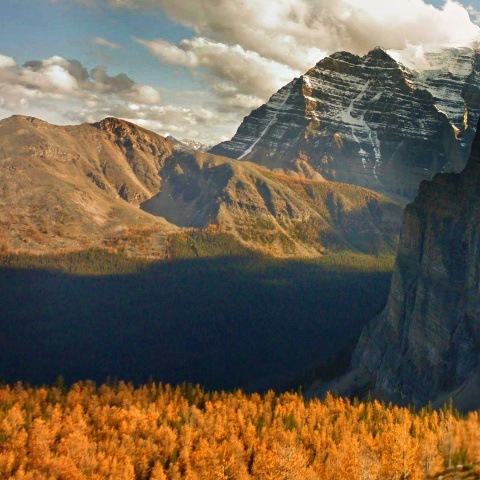}
\caption*{Predict: 2.7, Pro average: 3.3}
\end{subfigure}
\begin{subfigure}{.65\textwidth}
\centering
\includegraphics[width=1\linewidth]{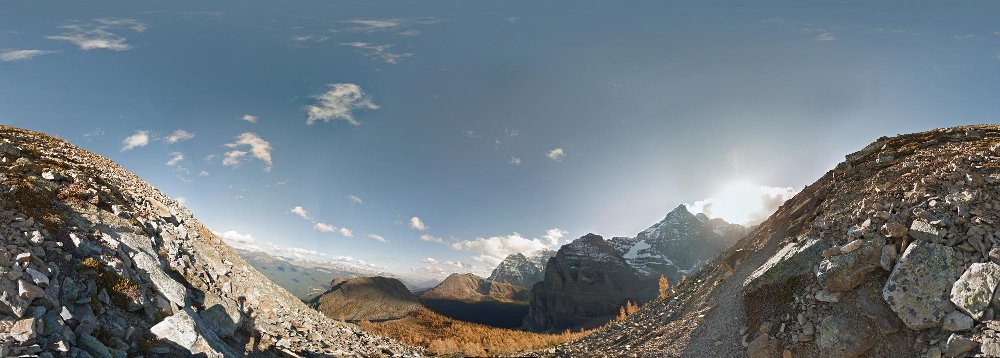}
\caption*{}
\end{subfigure} \\
\caption{Successful cases in our creation, with predicted and average professional rating.}
\label{fig:good_cases}
\end{figure*}

More successful cases are manually selected from all of the results,
and presented in Figure~\ref{fig:good_cases}.
Photos in left column are created from panoramas in right column.
For each photo, predicted score and average score from professionals are displayed.
The chance to encounter results with similar scores can be looked up in Figure~\ref{fig:pro_scores} (b).

We compiled a show-case webpage that contains additional results with nearby Street View panoramas at:
\href{https://google.github.io/creatism}{https://google.github.io/creatism}

\paragraph{Acknowledgement} The authors would like to acknowledge Vahid Kazemi for his earlier work
in predicting AVA ranking scores using Inception network, and Sagarika Chalasani, Nick Beato, 
Bryan Klingner and Rupert Breheny for their help in processing Google Street View panoramas.
We would like to thank Peyman Milanfar, Tom\'{a}\v{s} I\v{z}o, Christian Szegedy, Jon Barron and Sergey Ioffe
for their helpful comments. 
Huge thanks to our anonymous professional photographers. We hope you are not reading this paper to 
introduce bias in your future ratings!

{\small
\bibliography{main}

\begin{thebibliography}{}

\bibitem[Chen et~al., 2017]{DBLP:journals/corr/ChenKSCM17}
Chen, Y., Klopp, J., Sun, M., Chien, S., and Ma, K. (2017).
\newblock Learning to compose with professional photographs on the web.
\newblock {\em CoRR}, abs/1702.00503.

\bibitem[Gatys et~al., 2015]{DBLP:journals/corr/GatysEB15a}
Gatys, L.~A., Ecker, A.~S., and Bethge, M. (2015).
\newblock A neural algorithm of artistic style.
\newblock {\em CoRR}, abs/1508.06576.

\bibitem[Goodfellow et~al., 2014]{NIPS2014_5423}
Goodfellow, I., Pouget-Abadie, J., Mirza, M., Xu, B., Warde-Farley, D., Ozair,
  S., Courville, A., and Bengio, Y. (2014).
\newblock Generative adversarial nets.
\newblock In Ghahramani, Z., Welling, M., Cortes, C., Lawrence, N.~D., and
  Weinberger, K.~Q., editors, {\em Advances in Neural Information Processing
  Systems 27}, pages 2672--2680. Curran Associates, Inc.

\bibitem[Hertzmann et~al., 2001]{Hertzmann:2001:IA:383259.383295}
Hertzmann, A., Jacobs, C.~E., Oliver, N., Curless, B., and Salesin, D.~H.
  (2001).
\newblock Image analogies.
\newblock In {\em Proceedings of the 28th Annual Conference on Computer
  Graphics and Interactive Techniques}, SIGGRAPH '01, pages 327--340, New York,
  NY, USA. ACM.

\bibitem[Isola et~al., 2016]{DBLP:journals/corr/IsolaZZE16}
Isola, P., Zhu, J., Zhou, T., and Efros, A.~A. (2016).
\newblock Image-to-image translation with conditional adversarial networks.
\newblock {\em CoRR}, abs/1611.07004.

\bibitem[Jaroensri et~al., 2015]{acceptable}
Jaroensri, R., Paris, S., Hertzmann, A., Bychkovsky, V., and Durand, F. (2015).
\newblock Predicting range of acceptable photographic tonal adjustments.
\newblock In {\em IEEE International Conference on Computational Photography},
  Houston, TX.

\bibitem[Kong et~al., 2016]{DBLP:journals/corr/KongSLMF16}
Kong, S., Shen, X., Lin, Z., Mech, R., and Fowlkes, C.~C. (2016).
\newblock Photo aesthetics ranking network with attributes and content
  adaptation.
\newblock {\em CoRR}, abs/1606.01621.

\bibitem[Kopf et~al., 2007]{KCLU07}
Kopf, J., Cohen, M.~F., Lischinski, D., and Uyttendaele, M. (2007).
\newblock Joint bilateral upsampling.
\newblock {\em ACM Transactions on Graphics (Proceedings of SIGGRAPH 2007)},
  26(3):to appear.

\bibitem[Lu et~al., 2014]{Lu:2014:RRP:2647868.2654927}
Lu, X., Lin, Z., Jin, H., Yang, J., and Wang, J.~Z. (2014).
\newblock Rapid: Rating pictorial aesthetics using deep learning.
\newblock In {\em Proceedings of the 22Nd ACM International Conference on
  Multimedia}, MM '14, pages 457--466, New York, NY, USA. ACM.

\bibitem[Lu et~al., 2015a]{7243357}
Lu, X., Lin, Z., Jin, H., Yang, J., and Wang, J.~Z. (2015a).
\newblock Rating image aesthetics using deep learning.
\newblock {\em IEEE Transactions on Multimedia}, 17(11):2021--2034.

\bibitem[Lu et~al., 2015b]{7410476}
Lu, X., Lin, Z., Shen, X., Mech, R., and Wang, J.~Z. (2015b).
\newblock Deep multi-patch aggregation network for image style, aesthetics, and
  quality estimation.
\newblock In {\em 2015 IEEE International Conference on Computer Vision
  (ICCV)}, pages 990--998.

\bibitem[Luan et~al., 2017]{DBLP:journals/corr/LuanPSB17}
Luan, F., Paris, S., Shechtman, E., and Bala, K. (2017).
\newblock Deep photo style transfer.
\newblock {\em CoRR}, abs/1703.07511.

\bibitem[Mai et~al., 2016]{7780429}
Mai, L., Jin, H., and Liu, F. (2016).
\newblock Composition-preserving deep photo aesthetics assessment.
\newblock In {\em 2016 IEEE Conference on Computer Vision and Pattern
  Recognition (CVPR)}, pages 497--506.

\bibitem[Mantiuk et~al., 2012]{Mantiuk:2012:CFS:2393476.2393485}
Mantiuk, R.~K., Tomaszewska, A., and Mantiuk, R. (2012).
\newblock Comparison of four subjective methods for image quality assessment.
\newblock {\em Comput. Graph. Forum}, 31(8):2478--2491.

\bibitem[Marchesotti et~al., 2014]{DBLP:journals/corr/MarchesottiMP14}
Marchesotti, L., Murray, N., and Perronnin, F. (2014).
\newblock Discovering beautiful attributes for aesthetic image analysis.
\newblock {\em CoRR}, abs/1412.4940.

\bibitem[Mirza and Osindero, 2014]{DBLP:journals/corr/MirzaO14}
Mirza, M. and Osindero, S. (2014).
\newblock Conditional generative adversarial nets.
\newblock {\em CoRR}, abs/1411.1784.

\bibitem[Murray et~al., 2012]{6247954}
Murray, N., Marchesotti, L., and Perronnin, F. (2012).
\newblock Ava: A large-scale database for aesthetic visual analysis.
\newblock In {\em 2012 IEEE Conference on Computer Vision and Pattern
  Recognition}, pages 2408--2415.

\bibitem[Nguyen et~al., 2016]{DBLP:journals/corr/NguyenYBDC16}
Nguyen, A., Yosinski, J., Bengio, Y., Dosovitskiy, A., and Clune, J. (2016).
\newblock Plug {\&} play generative networks: Conditional iterative generation
  of images in latent space.
\newblock {\em CoRR}, abs/1612.00005.

\bibitem[Radford et~al., 2015]{DBLP:journals/corr/RadfordMC15}
Radford, A., Metz, L., and Chintala, S. (2015).
\newblock Unsupervised representation learning with deep convolutional
  generative adversarial networks.
\newblock {\em CoRR}, abs/1511.06434.

\bibitem[Szegedy et~al., 2015]{DBLP:journals/corr/SzegedyVISW15}
Szegedy, C., Vanhoucke, V., Ioffe, S., Shlens, J., and Wojna, Z. (2015).
\newblock Rethinking the inception architecture for computer vision.
\newblock {\em CoRR}, abs/1512.00567.

\bibitem[Yan et~al., 2016]{YanZW+16}
Yan, Z., Zhang, H., Wang, B., Paris, S., and Yu, Y. (2016).
\newblock Automatic photo adjustment using deep neural networks.
\newblock {\em {ACM} Transactions on Graphics}, 35(2):11.

\bibitem[Zhu et~al., 2017]{DBLP:journals/corr/ZhuPIE17}
Zhu, J., Park, T., Isola, P., and Efros, A.~A. (2017).
\newblock Unpaired image-to-image translation using cycle-consistent
  adversarial networks.
\newblock {\em CoRR}, abs/1703.10593.

\end{thebibliography}
\bibliographystyle{apalike}
}

\begin{table*}
\scriptsize 
\begin{tabular}{m{0.6cm}| m{16cm}}
\hline
\makecell{\textbf{Score}} & \makecell{\textbf{{Description}}} \\
\hline
1.0 & \textbf{Beginner: Point-and-shoot without consideration for composition, lighting etc. (it can still be technically good, i.e., good focus, good exposure...)}\\
& - This photo has no attention to light, composition, or moment and there is no clear intent. Looks like a photo taken by accident or like someone just held the camera up and pressed the shutter button... it's not clear why anyone would take this photo or want to look at this photo.\\ 
& - This score represents failure to use the camera properly and/or create anything that resembles a passable photo. \\
& - These photographs clearly are point and shoot with no intention of a composition in mind. These photographs usually lack contrast or thought and have nothing interesting in their subject matter. \\
& - No artistic merit or intention and does not follow basic rules of composition, lighting or focus.\\
\hline
2.0 &
\textbf{Amateur: Good photo for people without a background in photography. Nothing stands out as embarrassing. But nothing artistic either. Lighting and composition are OK. The majority of population falls between 1 and 2.}\\
& - I'm simply bored when I look at this image and it creates a sense of indifference. The image does not engage me at all and there is nothing interesting in the photo. This could be a combination of a poorly-lit image of a boring, cliche´d subject, and/or, an unsettled composition with poor tonal values. \\ 
& - This photo has a clear subject and things like composition and moment are ok but not great. You can tell that this person saw something of interest and they documented it, but they are still not thinking like a photographer because there is no attention to light, which is what photography is all about.\\ 
& - Pure point-and-shoot. These photos show us what is in front of the person taking the photograph. There is a passing sense for composing the elements within the frame. The angle is almost always at standing eye level. However, many of these images are acceptable to look at, and will often show interesting landscapes. The key point of classification here, is that the image does not exhibit professional photographic skill. A point-and-shoot photographer can still make a nice image if they are standing in front of beautiful landscape. \\
& - These photographs have some degree of thought put into the composition, with a strong image in mind. The image may have a clear message, but they miss the mark by either forgetting about contrast or composition. These are images that would usually fall into a camera phone photograph category.\\
& - Intention in composition, lighting or framing, but poorly executed.  Possible bad editing, out of focus, fragmenting, pixelation or poor quality.\\
\hline
2.5 &
\textbf{Something artistic obviously presents in this photo. However, the attempt in this photo, content or editing, can not be called successful.}\\
& - The photo feels like it was “taken”, not made. Though the image is clear, there may be an attempt at composition without resolve, there is no point of focus, and/or the tonal range may be only in the mid-tones making it “flat”. The light has not been considered to best capture the subject, and/or there are distracting objects in the photo that keep it from the 3, 3.5, and 4 caliber. \\ 
& - You can tell in this photo, that the person is paying attention to light! They are starting to think like a photographer. Things like composition, moment and subject are an improvement but still lacking. There is attention to the way that light interacts with a subject or environment, even though the usage of light might not be very good.\\
& - This rating marks a step in the right direction beyond point-and-shoot photography. There is an effort here to create a better photo (interesting angle, composition, use of silhouette, compelling lighting, etc....) However, the image still does not fully add up to a well-made photograph. (These are the sort of images you'll see in a Photo-1 class. Good effort and intentions, but more skill needs to be applied). \\
& - These photographs have the intention of a good photograph but are missing out on many of the key elements of a professional photograph. This usually means lack of focus or composition. The photo may have a beautiful image but is cropping out a person. The photograph may have a beautiful mountainside, but the entire image is not straight. \\
& - Average photo, not good or bad.  Follows rules of composition with lighting and framing, but not particularly well executed.\\
\hline
3.0 &
\textbf{Semi-Pro: One is on a path to become a professional photographer!}\\
& - I feel this is the critical break-point. The image is “good” with an effort to make/capture the shot, but the photo reveals a skill-level of one who does not have a lot of experience making great images and may employ the use of Photoshop in an attempt to enhance the photo. This is what I call “over-cooking” the image. It cheapens the photo and is a dead giveaway of an exploring ameteur. It is almost as if the photographer is trying too hard. The photo may also be technically and aesthetically missing an element that would make it a 3.5. I would expect to find many of this caliber of imagery in a local art fair. \\ 
& - This photo has attention to light, a clear subject, good composition and a clear intent but more than a few factors are still lacking. Usage of light is better, but not great. the moment is a little awkward and the subject is boring.\\
& - This is a good photograph that works. The general approach has created a worthwhile landscape. This often includes at least one professional strategy that brings the image together (strong composition, depth-of-field, interesting angle, compelling lighting, etc....) \\
& - These photographs show a strong understanding of imagery and composition with a clear intention. These photographs fall short when it comes to a subject matter that defines a perfect photograph. \\
& - Above average image with clear thought, focus and framing put into it.\\
\hline
3.5 &
\\
& - The image is better than most, but has been done before in a more complete way. Usually the subject is amazing but the lighting could be better at a different time of day, or has any combination of great and slightly sub-great components. It’s almost a 4, but I reserve 4’s for only the best. \\ 
& - This photo has excellent use of light, a clear subject, a clear intent and almost all of the characteristics of a professionally crafted photo but there is just one factor that's off. Either the subject is boring, the moment is a little awkward, or the composition is a little messy. \\ 
& - This can be a tricky rating. For this, I often ask "what could have been done here that would make this photo even better, and worthy of a 4 rating?" In that sense, I use 3.5 to mark down from 4. Maybe the photographer oversaturated a perfectly good landscape, maybe they collided some elements within the frame (Ansel Adams often mentioned this). Or maybe the image just needs one more element, something that a professional would be mindful of. \\
& - These photographs have beautiful imagery but do not have the focus or the perfect composition that make a photograph truly professional. These are usually almost perfect photographs but are missing out on the techniques that make a photograph stand out as a perfect image.\\
& - Great image with purposeful depth of field and framing clearly taken by someone with photographic knowledge.\\
\hline
4.0 &
\textbf{Pro: photos you think deserved to be called taken by a professional.}\\
& - This photo was made, not taken. Everything in the image is working together to the sum of a great image. Without question this image was made by a skilled craftsman, one who is technically fluent, environmentally aware, has good timing and/or patience, is in command of post production and does not use clichéd and overused filters, and offers a controlled composition that has a relationship with the subject. Anyone who sees this image would consider it “professional”. \\ 
& - This photo was created by someone who has studied photography and refined their craft. There is a great moment/interesting subject in great light. There is meaningful interaction between light and subject. Excellent use of composition. The moment is just right, and you can clearly see the photographer's intent. \\ 
& - This is a well-made professional photograph which exhibits experience, technical know-how, and above all else - a sense for the strategies which go into making strong landscape imagery. \\
& - These photographs are clearly shot by a professional with a precise composition in mind. There is a strong contrast of darks and lights. These photographs use techniques that show a strong understanding of their camera equipment.\\
& - Excellent image, reserved only for the best images with well thought out intentional and dynamic compositions, good lighting with balance in colors and tones and purposefully in or out of focus.\\
\hline
\end{tabular}
\caption{For each aesthetic score, text in bold summarizes our initial description, followed by selected comments
from professional photographers.}
\vskip -1em
\label{table:scores_feedback} 
\end{table*}

\begin{figure*}[t]
\begin{subfigure}{.40\textwidth}
\centering
\includegraphics[width=1\linewidth]{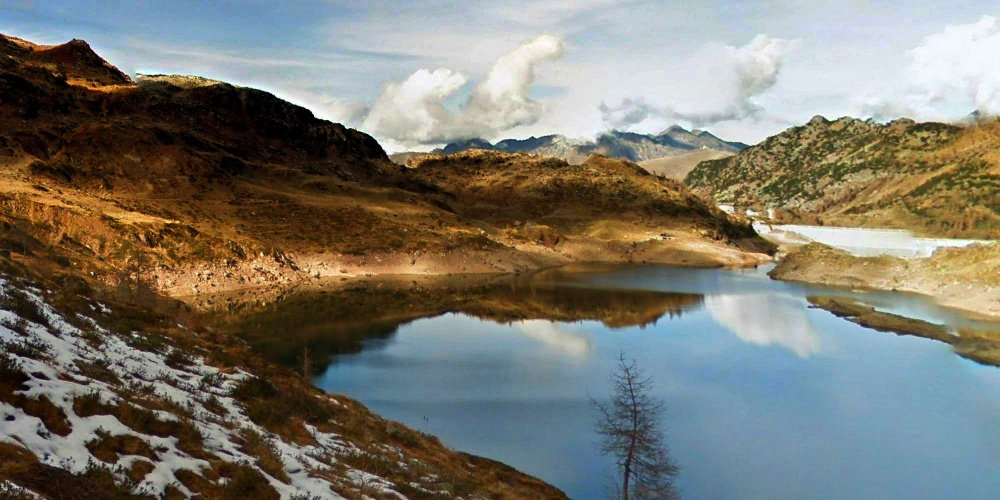}
\caption*{Predict: 2.6, Pro average: 3.3}
\end{subfigure}
\begin{subfigure}{.555\textwidth}
\centering
\includegraphics[width=1\linewidth]{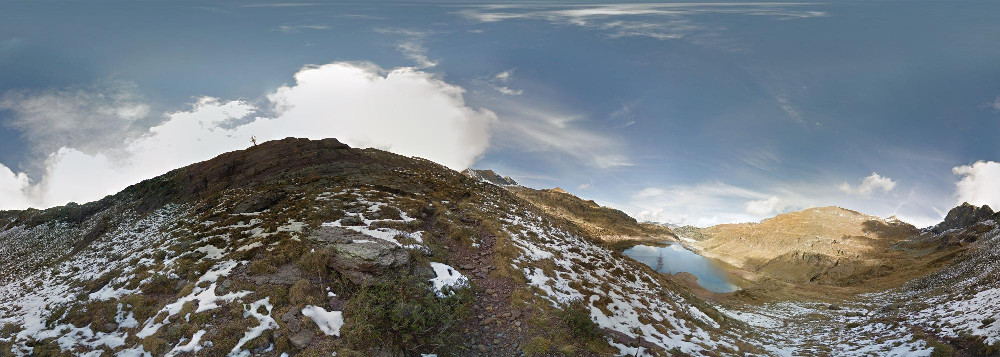}
\caption*{}
\end{subfigure} \\
\begin{subfigure}{.40\textwidth}
\centering
\includegraphics[width=1\linewidth]{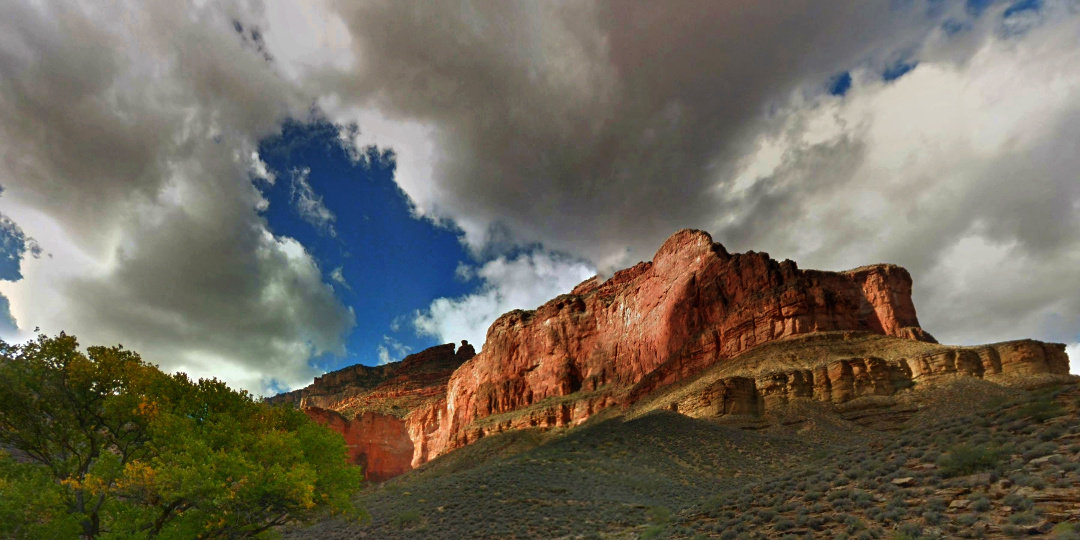}
\caption*{Predict: 2.7, Pro average: 3.0}
\end{subfigure}
\begin{subfigure}{.555\textwidth}
\centering
\includegraphics[width=1\linewidth]{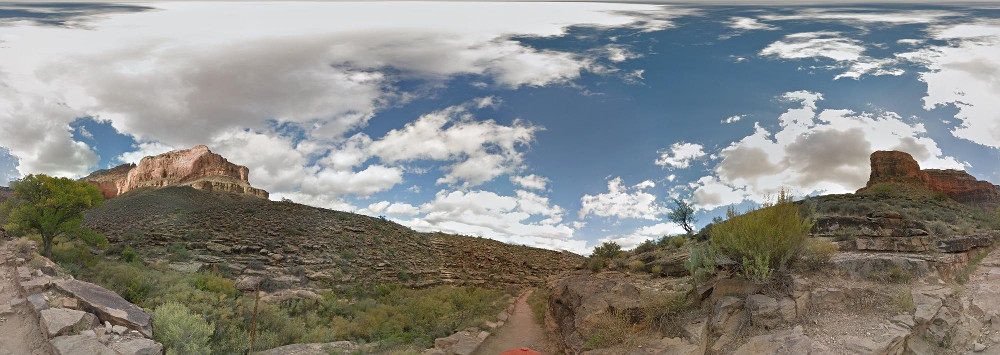}
\caption*{}
\end{subfigure} \\
\begin{subfigure}{.40\textwidth}
\centering
\includegraphics[width=1\linewidth]{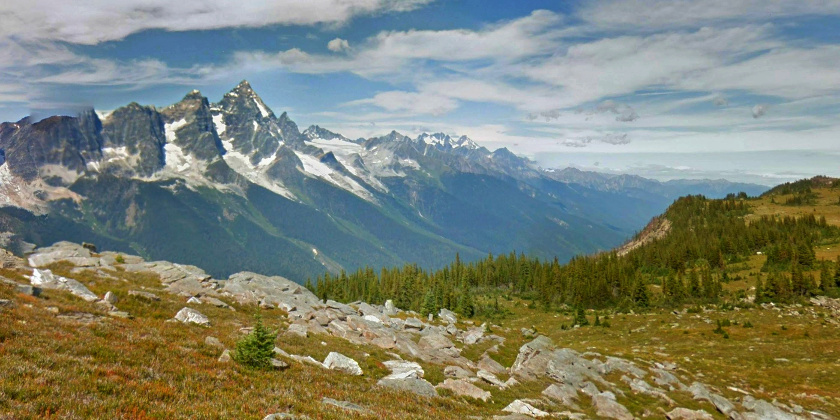}
\caption*{Predict: 2.6, Pro average: 3.0}
\end{subfigure}
\begin{subfigure}{.555\textwidth}
\centering
\includegraphics[width=1\linewidth]{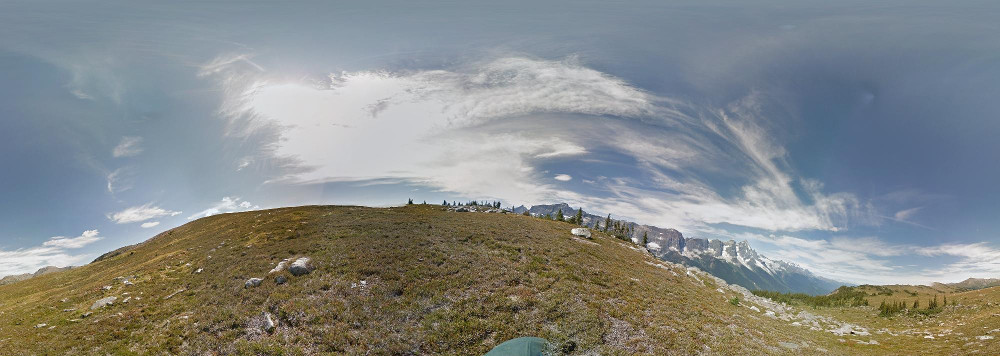}
\caption*{}
\end{subfigure} \\
\begin{subfigure}{.40\textwidth}
\centering
\includegraphics[width=1\linewidth]{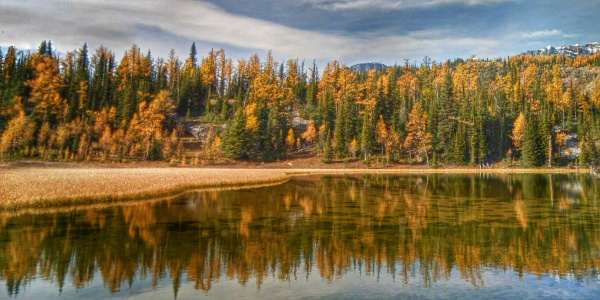}
\caption*{Predict: 2.6, Pro average: 3.3}
\end{subfigure}
\begin{subfigure}{.555\textwidth}
\centering
\includegraphics[width=1\linewidth]{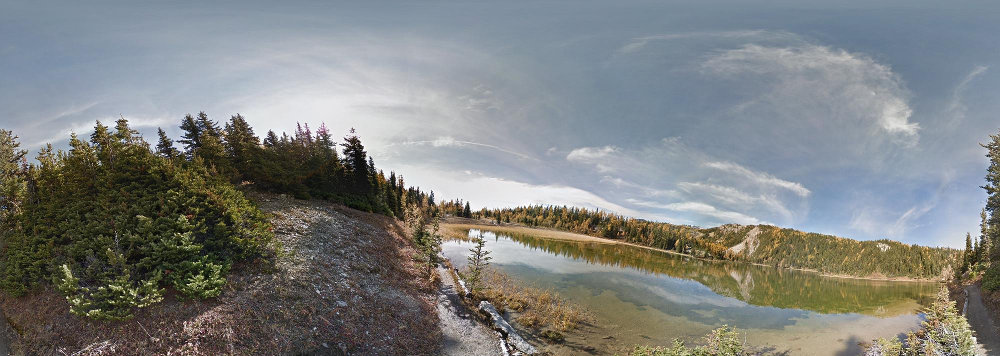}
\caption*{}
\end{subfigure} \\
\begin{subfigure}{.40\textwidth}
\centering
\includegraphics[width=1\linewidth]{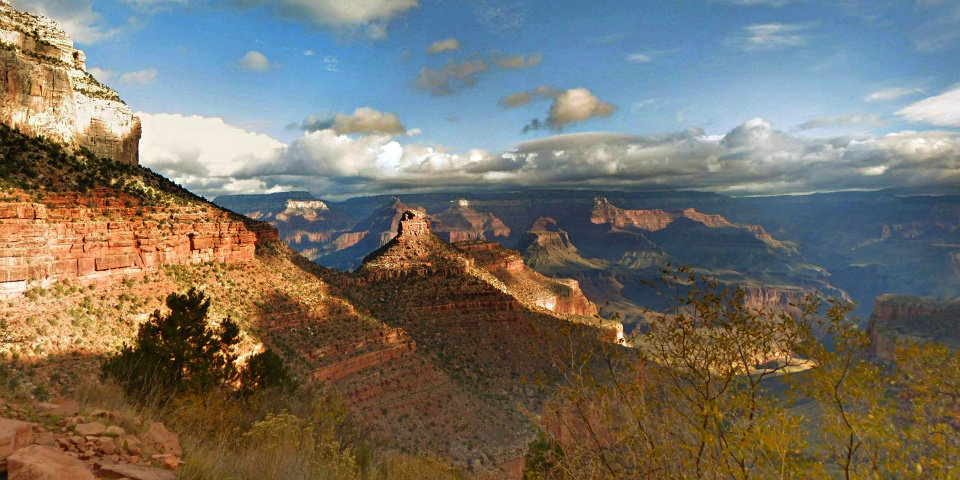}
\caption*{Predict: 2.8, Pro average: 3.3}
\end{subfigure}
\begin{subfigure}{.555\textwidth}
\centering
\includegraphics[width=1\linewidth]{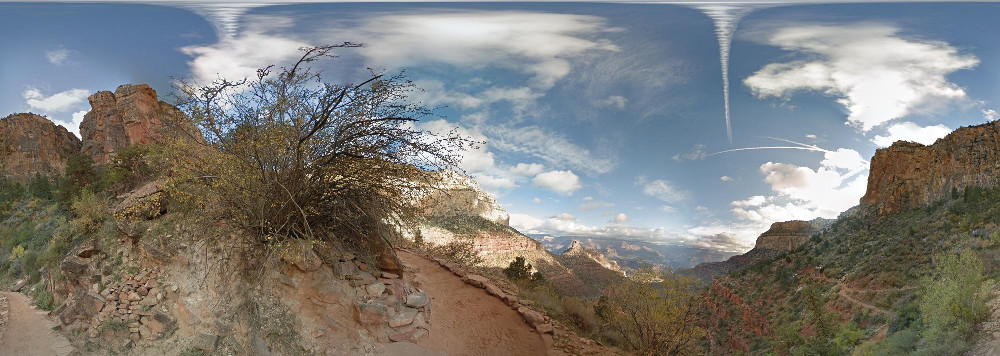}
\caption*{}
\end{subfigure} \\
\begin{subfigure}{.40\textwidth}
\centering
\includegraphics[width=1\linewidth]{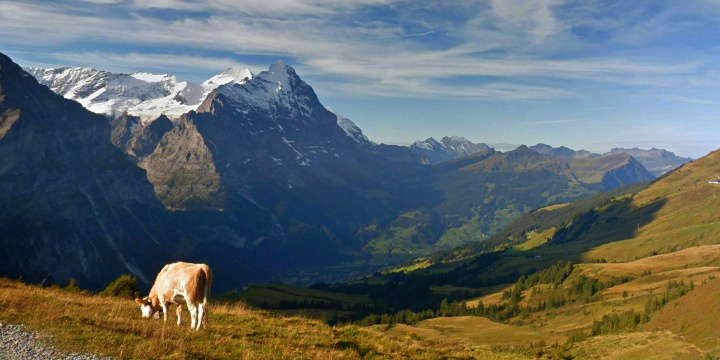}
\caption*{Predict: 2.8, Pro average: 3.2}
\end{subfigure}
\begin{subfigure}{.555\textwidth}
\centering
\includegraphics[width=1\linewidth]{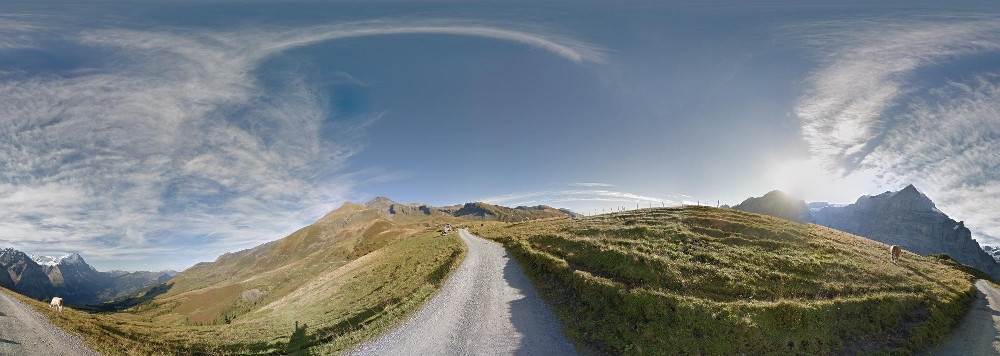}
\caption*{}
\end{subfigure} \\
\caption*{}
\end{figure*}

\begin{figure*}[t]
\begin{subfigure}{.40\textwidth}
\centering
\includegraphics[width=1\linewidth]{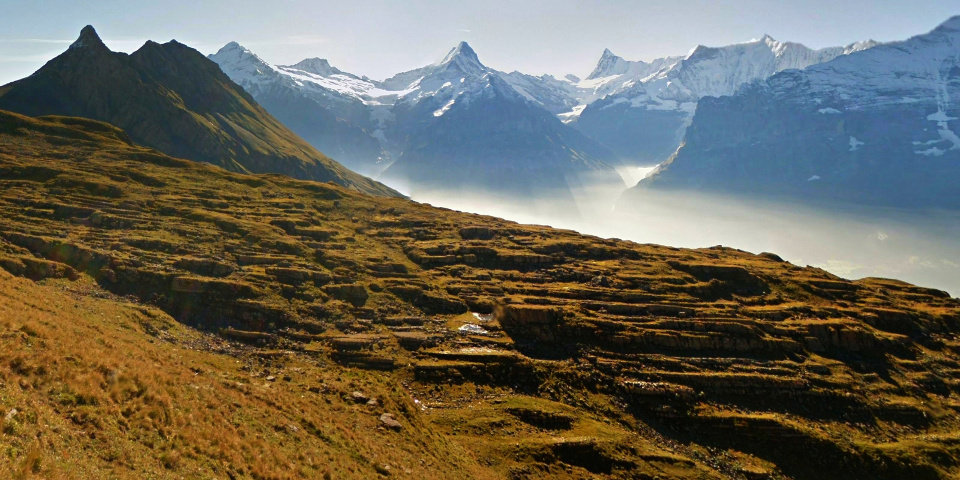}
\caption*{Predict: 2.8, Pro average: 2.8}
\end{subfigure}
\begin{subfigure}{.555\textwidth}
\centering
\includegraphics[width=1\linewidth]{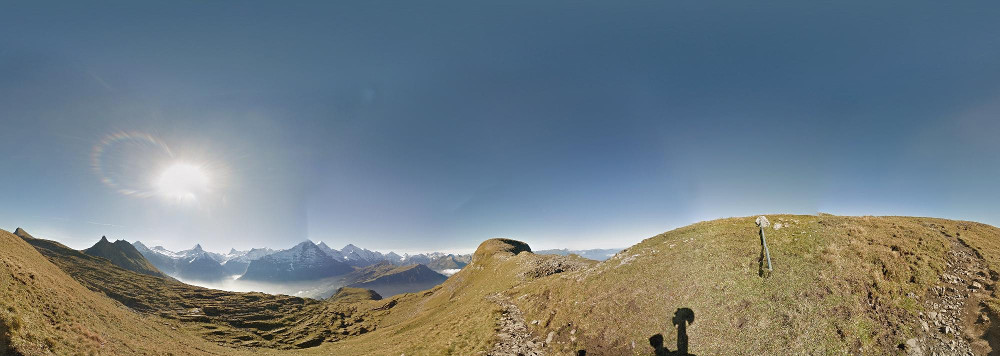}
\caption*{}
\end{subfigure} \\
\begin{subfigure}{.40\textwidth}
\centering
\includegraphics[width=1\linewidth]{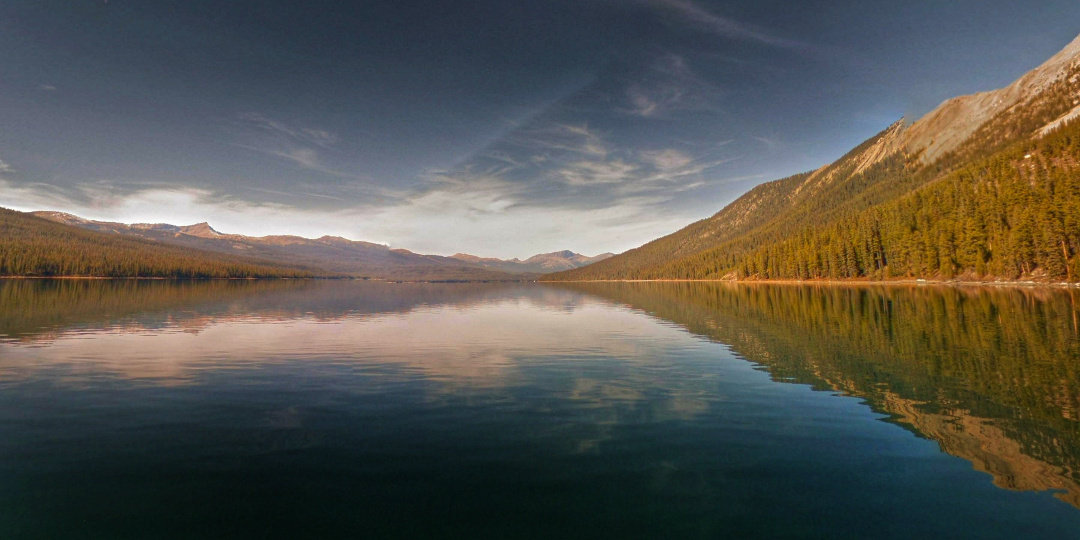}
\caption*{Predict: 2.7, Pro average: 3.3}
\end{subfigure}
\begin{subfigure}{.555\textwidth}
\centering
\includegraphics[width=1\linewidth]{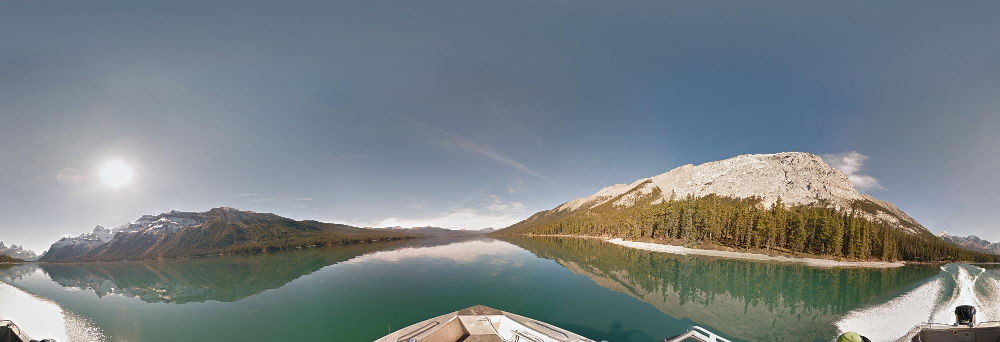}
\caption*{}
\end{subfigure} \\
\begin{subfigure}{.40\textwidth}
\centering
\includegraphics[width=1\linewidth]{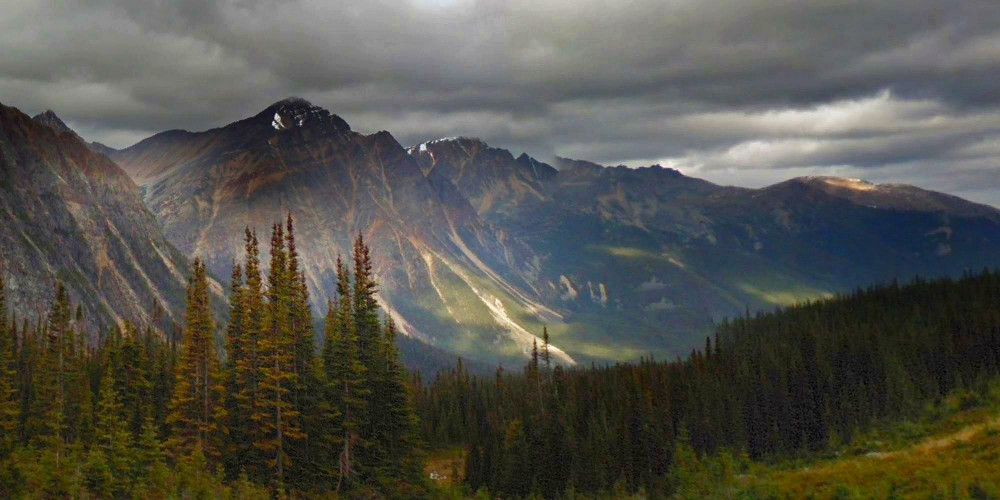}
\caption*{Predict: 2.6, Pro average: 3.5}
\end{subfigure}
\begin{subfigure}{.555\textwidth}
\centering
\includegraphics[width=1\linewidth]{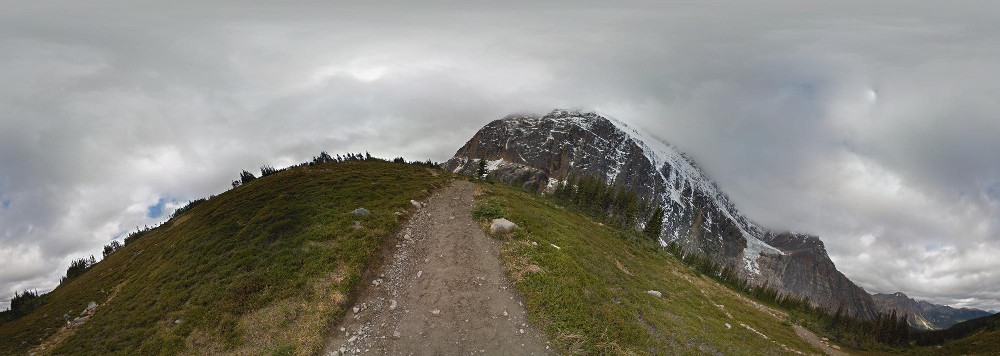}
\caption*{}
\end{subfigure} \\
\begin{subfigure}{.40\textwidth}
\centering
\includegraphics[width=1\linewidth]{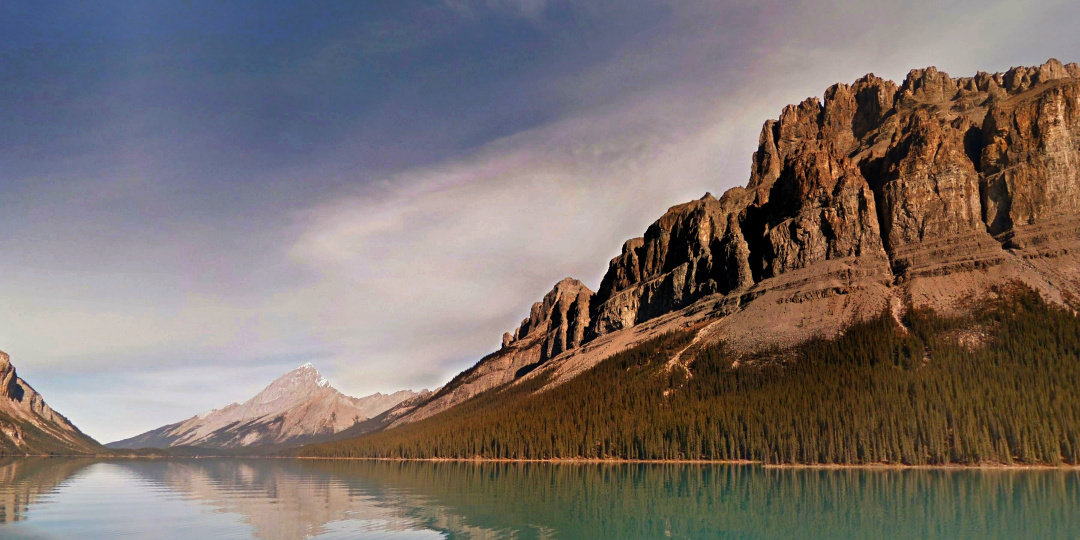}
\caption*{Predict: 2.9, Pro average: 3.3}
\end{subfigure}
\begin{subfigure}{.555\textwidth}
\centering
\includegraphics[width=1\linewidth]{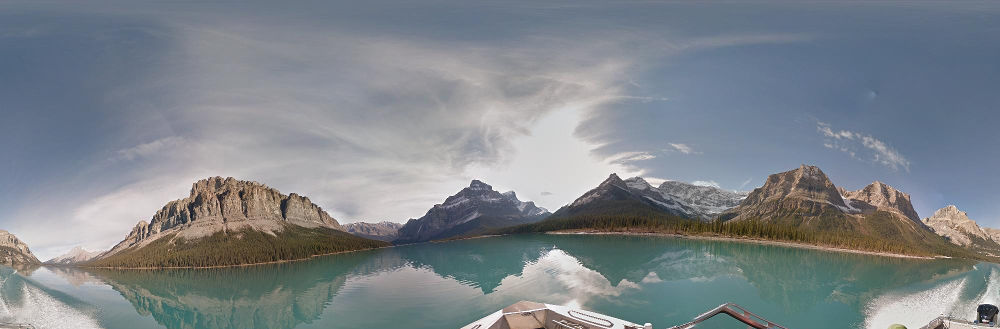}
\caption*{}
\end{subfigure} \\
\begin{subfigure}{.40\textwidth}
\centering
\includegraphics[width=1\linewidth]{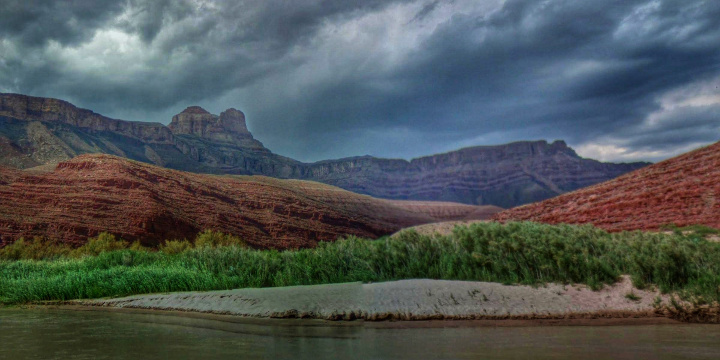}
\caption*{Predict: 2.4, Pro average: 2.8}
\end{subfigure}
\begin{subfigure}{.555\textwidth}
\centering
\includegraphics[width=1\linewidth]{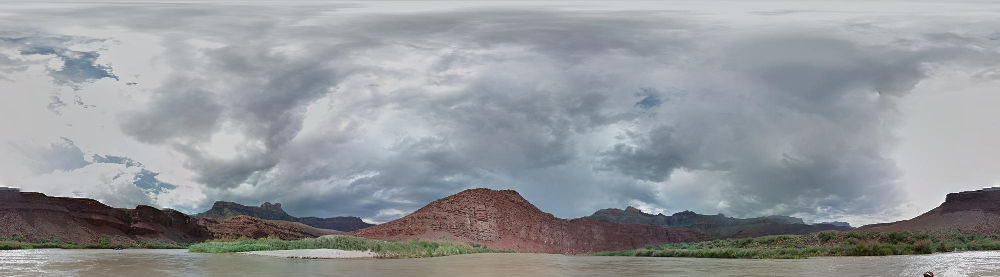}
\caption*{}
\end{subfigure} \\
\begin{subfigure}{.40\textwidth}
\centering
\includegraphics[width=1\linewidth]{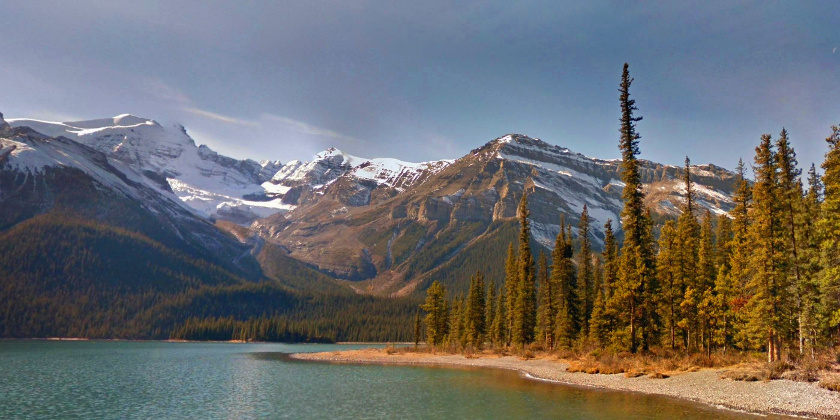}
\caption*{Predict: 2.8, Pro average: 3.3}
\end{subfigure} 
\begin{subfigure}{.555\textwidth}
\centering
\includegraphics[width=1\linewidth]{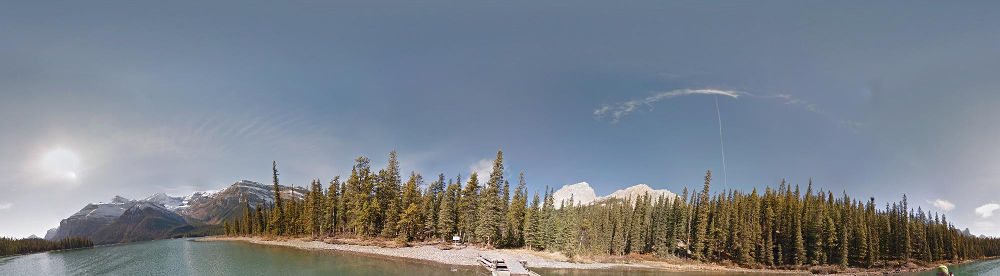}
\caption*{}
\end{subfigure} \\
\caption*{}
\end{figure*}

\begin{figure*}[t]
\begin{subfigure}{.33\textwidth}
\centering
\includegraphics[width=1\linewidth]{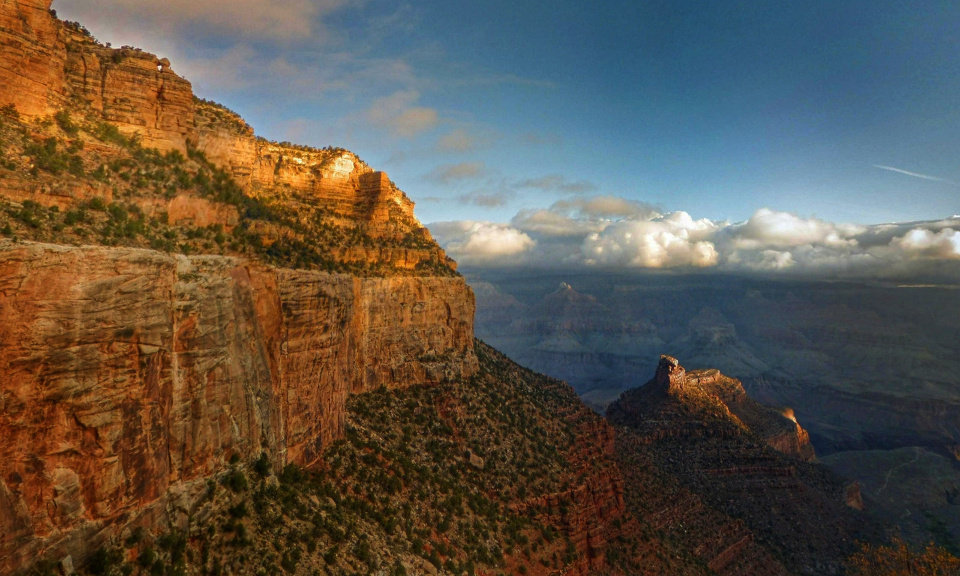}
\caption*{Predict: 2.8, Pro average: 3.0}
\end{subfigure}
\begin{subfigure}{.555\textwidth}
\centering
\includegraphics[width=1\linewidth]{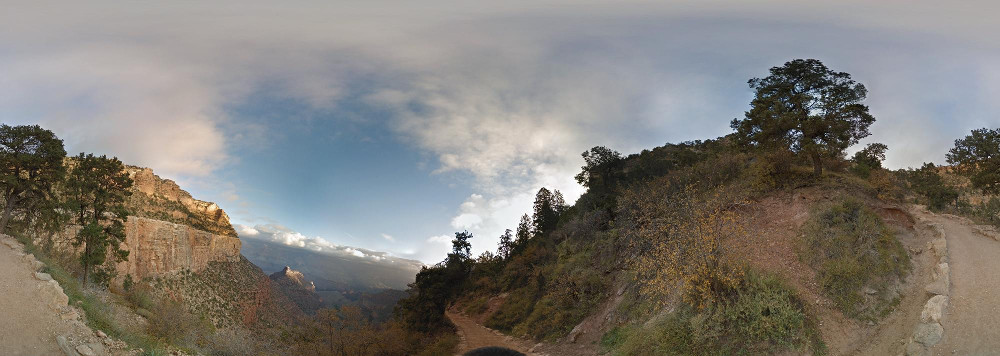}
\caption*{}
\end{subfigure} \\
\begin{subfigure}{.33\textwidth}
\centering
\includegraphics[width=1\linewidth]{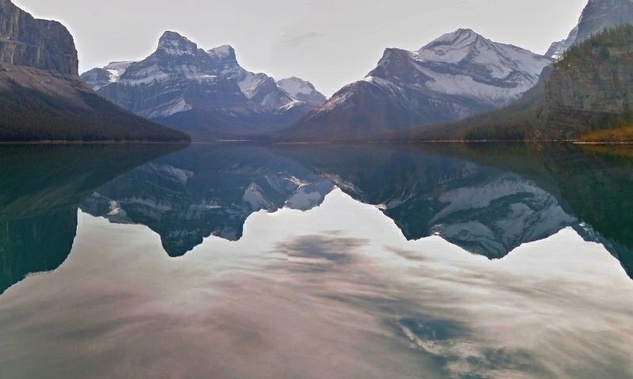}
\caption*{Predict: 2.9, Pro average: 3.8}
\end{subfigure}
\begin{subfigure}{.555\textwidth}
\centering
\includegraphics[width=1\linewidth]{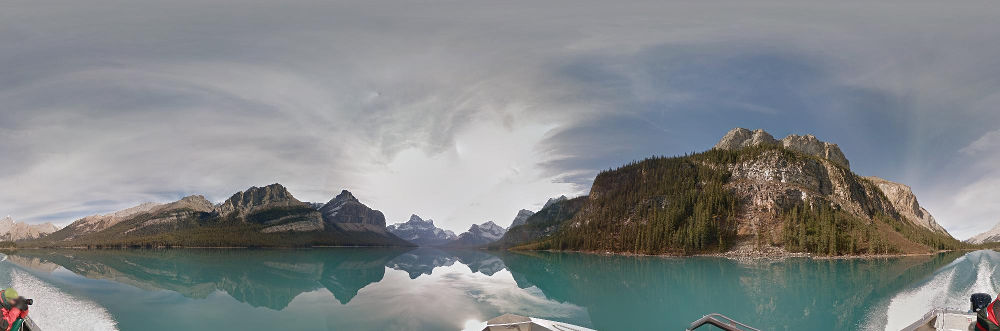}
\caption*{}
\end{subfigure} \\
\begin{subfigure}{.33\textwidth}
\centering
\includegraphics[width=1\linewidth]{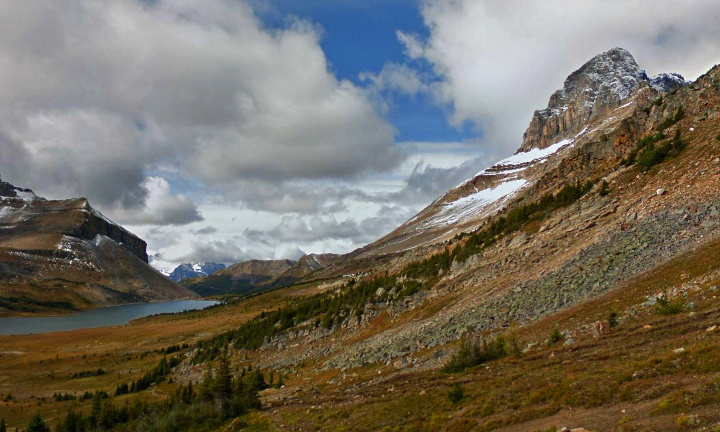}
\caption*{Predict: 2.9, Pro average: 2.8}
\end{subfigure}
\begin{subfigure}{.555\textwidth}
\centering
\includegraphics[width=1\linewidth]{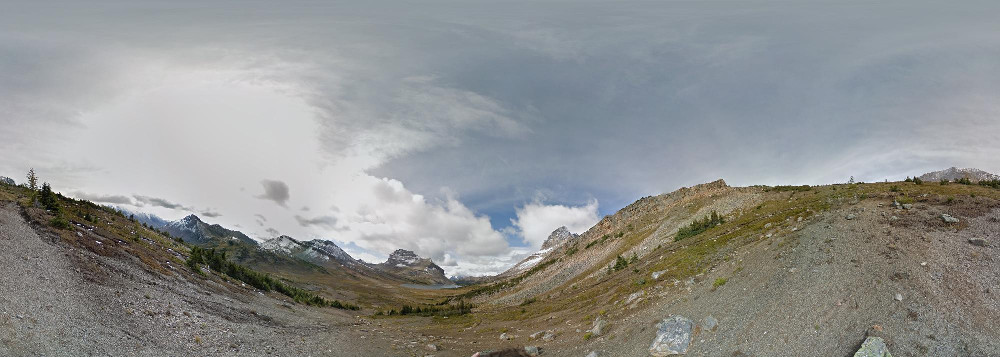}
\caption*{}
\end{subfigure} \\
\begin{subfigure}{.33\textwidth}
\centering
\includegraphics[width=1\linewidth]{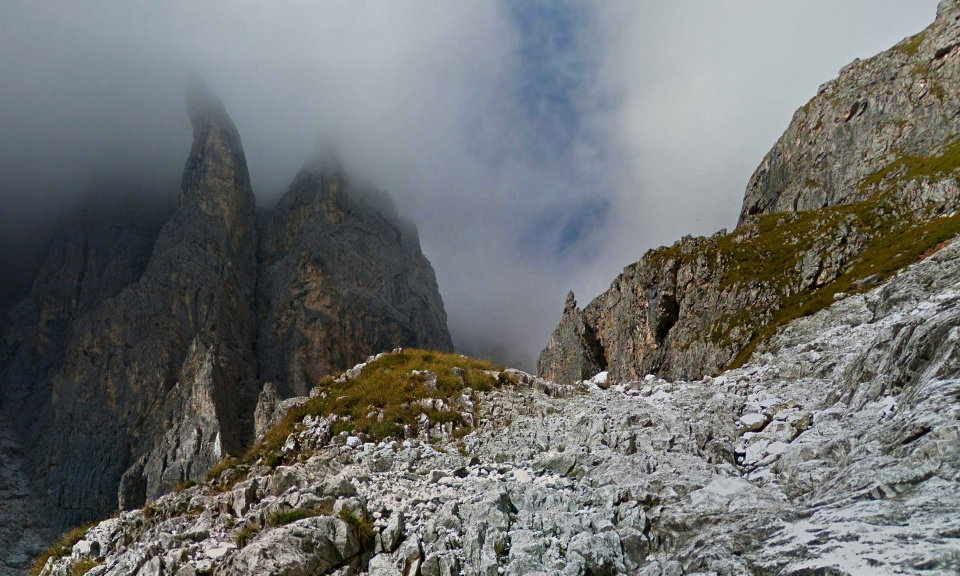}
\caption*{Predict: 2.4, Pro average: 2.7}
\end{subfigure}
\begin{subfigure}{.555\textwidth}
\centering
\includegraphics[width=1\linewidth]{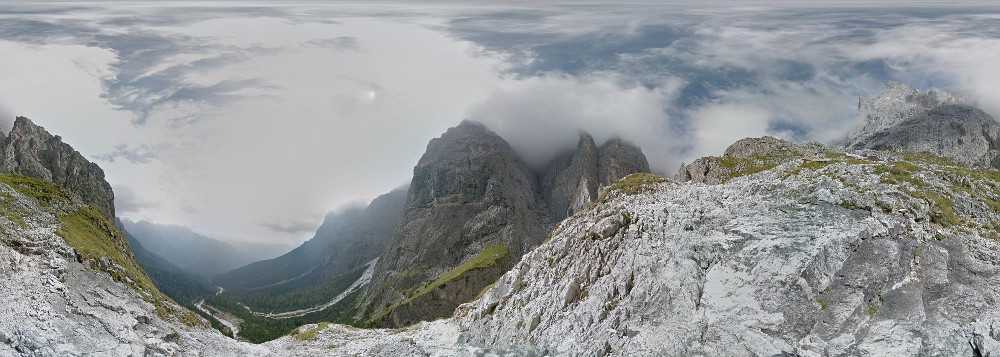}
\caption*{}
\end{subfigure} \\
\begin{subfigure}{.33\textwidth}
\centering
\includegraphics[width=1\linewidth]{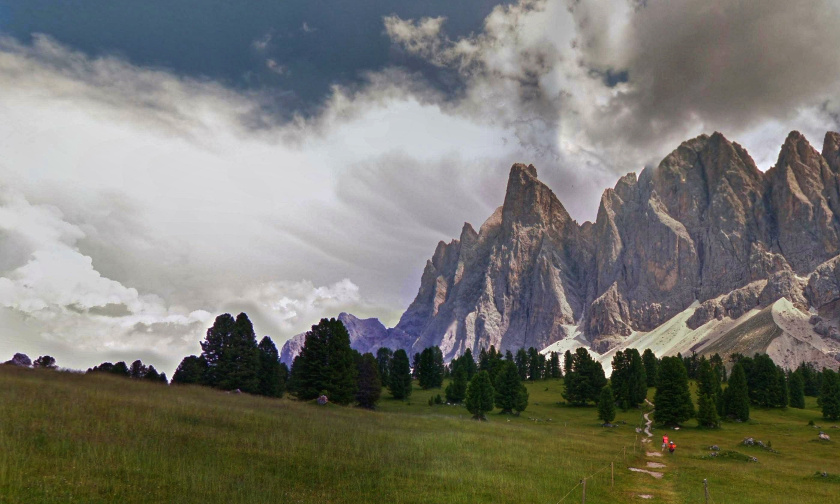}
\caption*{Predict: 2.2, Pro average: 3.0}
\end{subfigure}
\begin{subfigure}{.555\textwidth}
\centering
\includegraphics[width=1\linewidth]{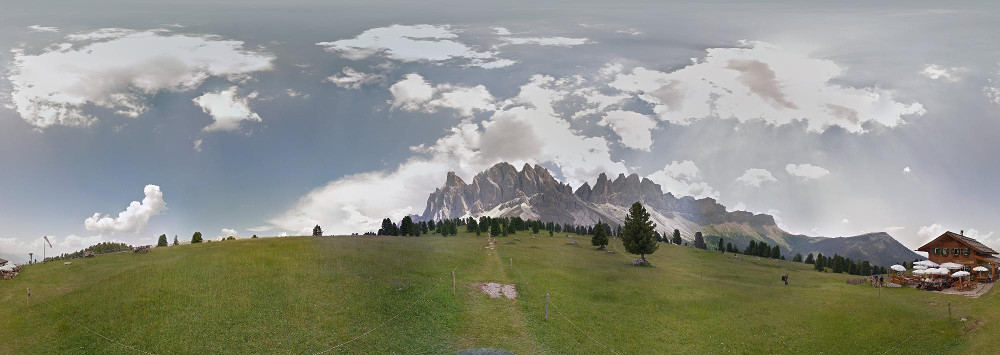}
\caption*{}
\end{subfigure} \\
\begin{subfigure}{.33\textwidth}
\centering
\includegraphics[width=1\linewidth]{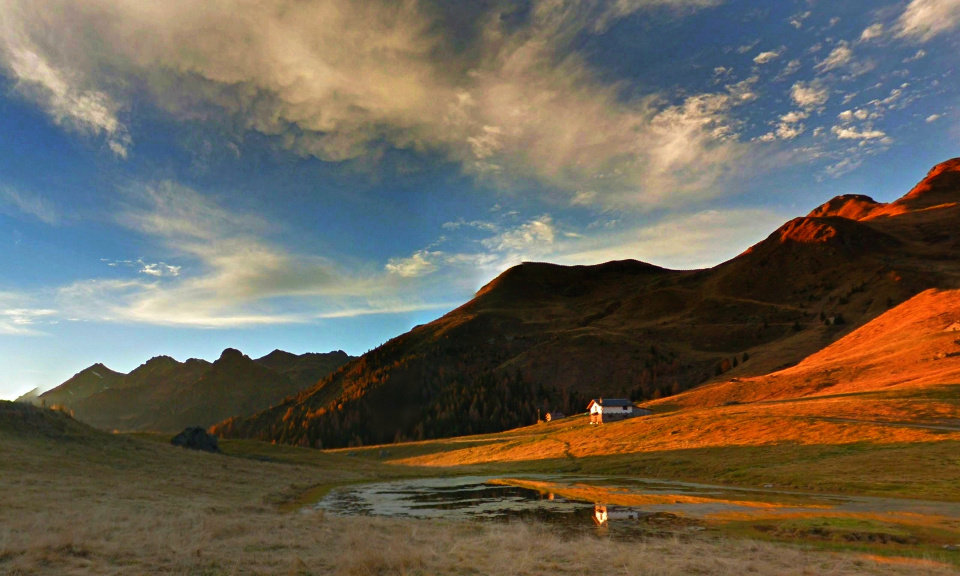}
\caption*{Predict: 2.8, Pro average: 3.5}
\end{subfigure}
\begin{subfigure}{.555\textwidth}
\centering
\includegraphics[width=1\linewidth]{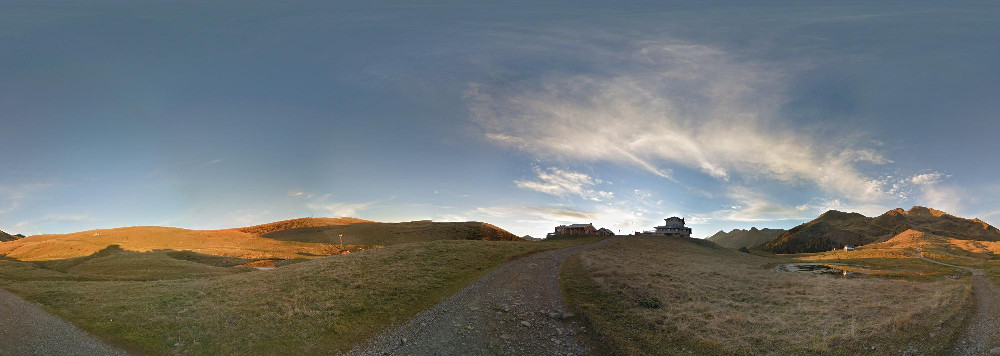}
\caption*{}
\end{subfigure} \\
\caption*{}
\end{figure*}

\end{document}